\documentclass{article} % For LaTeX2e
\usepackage{iclr2027_conference,times}

\usepackage{amsmath,amsfonts,bm}

\def\eqref#1{equation~\ref{#1}}
\def\1{\bm{1}}

\DeclareMathAlphabet{\mathsfit}{\encodingdefault}{\sfdefault}{m}{sl}
\SetMathAlphabet{\mathsfit}{bold}{\encodingdefault}{\sfdefault}{bx}{n}

\usepackage{hyperref}
\usepackage{url}
\usepackage{multirow}
\usepackage{amsmath}
\usepackage{amssymb}
\usepackage{booktabs}
\usepackage{graphicx}
\usepackage{cleveref}
\usepackage{capt-of}
\usepackage{xspace}
\usepackage{makecell}
\usepackage{subcaption}
\usepackage{wrapfig}
\usepackage{enumitem}
\usepackage{xspace}
\newcommand{\method}{LoGo\xspace}

\definecolor{darkblue}{rgb}{0, 0, 0.5}
\definecolor{darkred}{rgb}{0.6, 0, 0}
\hypersetup{colorlinks=true, citecolor=darkblue, linkcolor=darkred, urlcolor=darkblue}

\newcommand{\mypar}[1]{\vspace{1mm}\noindent\textbf{#1}}

\def\ie{\emph{i.e.,}\xspace}

\title{LoGo: Local-Global Rewards for Consistent Long-Horizon Video Generation}

\author{
\parbox{\textwidth}{\raggedright\bfseries
Ziqi Ma$^{1,2*}$, Shreya Sharma$^{2}$, Mohamed El Banani$^{2}$, Katja Schwarz$^{2}$, Chongjie Ye$^{2}$, Chao-Yuan Wu$^{2}$, Li Fei-Fei$^{2}$, Ben Mildenhall$^{2}$, Georgia Gkioxari$^{1}$, Justin Johnson$^{2}$, Gowthami Somepalli$^{2}$} \\
\\
$^{1}$California Institute of Technology \quad $^{2}$World Labs \\
{\small $^{*}$Work done during an internship at World Labs}
}

\iclrfinalcopy % Uncomment for camera-ready version, but NOT for submission.
\begin{document}

%\maketitle

{%
\maketitle
\begin{center}
    %\vspace{-6mm}
    \centering
    \includegraphics[width=0.95\linewidth]{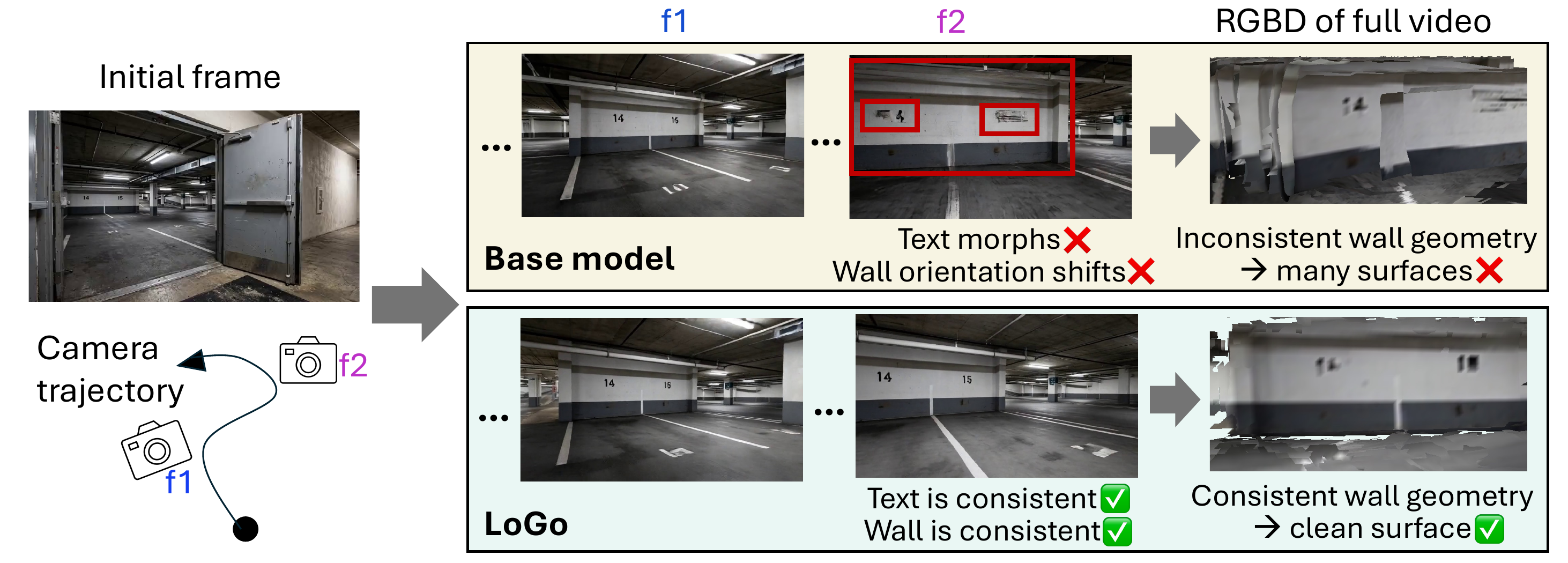}
    \vspace{-2mm}
    \captionof{figure}{Camera-controlled video models often break 3D consistency during generation (top row). Here, a video generated by an existing video model shows both local inconsistencies (the numbers on the wall are distorted) and global inconsistencies (the wall's orientation shifts).
    We introduce \method, a post-training approach that combines global and spatially localized rewards to reliably improve consistency in generation (bottom row).}
    \label{fig:teaser}
    \vspace{-1mm}
\end{center}
}

\begin{abstract}
Camera-controlled video models are rapidly advancing toward long generation horizons and complex camera control.
A key failure mode is 3D inconsistency: as the camera moves, objects lose permanence and scene structures shift.
Existing post-training techniques, which assign a single scalar reward to the entire generation, are poorly suited to correcting these inconsistencies over long horizons.
We introduce \method, which blends global and spatially localized rewards for camera-controlled video models. 
The local reward provides fine-grained credit assignment, which substantially improves 3D consistency, while the global reward preserves camera following and video quality.
Across three base models, \method shows a clear advantage on DL3DV and TrajectoryBench, a new benchmark for long-horizon, complex-camera-control generation that current evaluations lack. \method effectively reduces local object shifts, artifacts, and global scene changes, illustrating the importance of credit assignment in post-training video models. Project website: \url{https://ziqi-ma.github.io/logo-website/}
\end{abstract}    
\section{Introduction}
\label{sec:intro}

The rapid development in video generation has enabled us to generate \textit{worlds} from image and text prompts. Growing camera control capabilities enable users to freely navigate the generated space. But, what qualities are necessary to transform these generates frames into ``worlds'' that users can immerse themselves in and embodied applications can leverage?

% importance of consistency, it's a big challenge
A fundamental requirement is consistency: scenes should have persistent structure, and objects should have permanence. However, current models struggle; \cite{kupyn2025epipolar} highlights classical geometric errors, \cite{rakheja2025world} studies object and relation permanence, and \cite{ye2026mind} focuses on revisit consistency. These works highlight the difficulty of consistent video generation. In this work, we focus on 3D consistency: \textbf{(1)} Globally, the spatial structure and layout should persist as the user navigates the scene; \textbf{(2)} Locally, objects should stay consistent---they should not suddenly appear or disappear, become corrupted by artifacts, or otherwise shift in geometry or appearance.

\begin{figure*}[t]
\centering
\includegraphics[width=\textwidth]{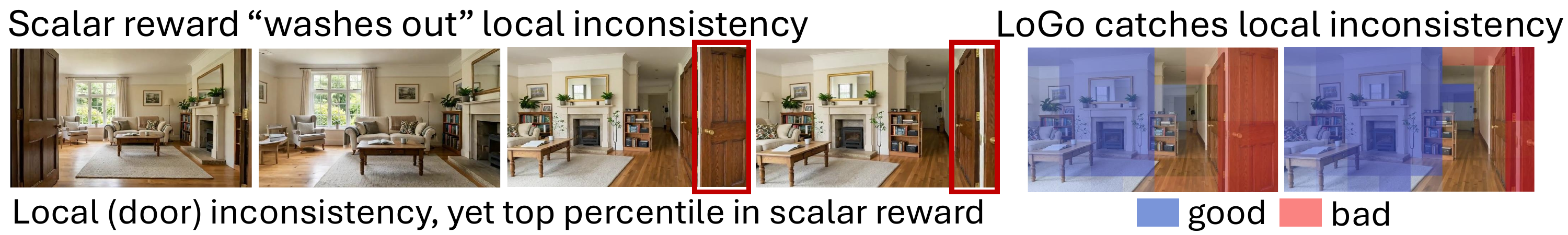}
\vspace{-6mm}
\caption{A scalar reward cannot capture localized failures: the door inconsistency is ``washed out'' by the overall good structure of the rollout. \method catches the localized inconsistency (right).}
\label{fig:motivation}
\vspace{-5mm}
\end{figure*}

% introduce existing methods
Current approaches improve video generation consistency through post-training. They apply DPO~\citep{rafailov2023direct} or Flow-GRPO~\citep{liu2026flow} with geometry rewards such as epipolar constraints~\citep{kupyn2025epipolar}, depth reprojection~\citep{an2026vggrpo, du2026videogpa}, or Gaussian reconstruction~\citep{wang2026world}. These approaches all collapse the reward to a single value for the entire generated video, without fine-grained credit assignment.
% why previous methods fail
This may suffice for short-horizon generation (80 frames), but new challenges emerge as horizons grow. Chief among them is 3D consistency failures, which tend to be localized and are thus difficult to capture with a single global reward. Such failures surface in our experiments generating up to 400 frames. For example, in Fig.~\ref{fig:motivation}, the door inconsistency in the rollout is local, yet the clip earns a high score for its otherwise sound geometry, and the error goes unpenalized.

%% -------- FIGURE -------- %%
%% ---------------- %%

% introduce our method
We propose \method, which enhances the global reward with a spatially localized reward. \method directly performs credit assignment in 3D space. As shown by the reward heatmap on the right of Fig.~\ref{fig:motivation}, \method pinpoints the door inconsistency, promoting local as well as global consistency. This local reward improves 3D consistency, and when combined with camera and aesthetic rewards it preserves or even improves video quality and camera following.

% benchmark
Current benchmarks for 3D consistency~\citep{duan2025worldscore} focus on short-horizon (80-frame) generations with simple camera commands like ``push forward.''
To address the lack of 3D consistency evaluation for long-horizon generation, we introduce TrajectoryBench, a new benchmark centered on complex camera trajectories. TrajectoryBench challenges models to generate space far beyond the initial view and to revisit the same regions from new angles, exposing failure modes that arise under these demanding conditions and giving the community a testbed to study them.

% results
Across three base video models, \method shows a strong advantage compared to baseline methods using global reward structure. \method's advantage persists across three different post-training algorithms, demonstrating the robustness of our method.
We make the following contributions: \textbf{(1)} We propose \method, a 3D-based reward structure that improves geometric consistency in camera-controlled video generation via combined global and spatially localized rewards, achieving state-of-the-art on two challenging benchmarks; \textbf{(2)} Our reward framework is general, yielding consistent gains across a range of base models -- Lingbot2~\citep{lingbot-world-v2}, Lyra2~\citep{shen2026lyra2}, and UniWorld~\citep{zhou2026uniworldview} -- and post-training techniques -- DiffusionNFT~\citep{zheng2026diffusionnft}, Flow-GRPO~\citep{liu2026flow}, and DRaFT~\citep{clark2024directly}; \textbf{(3)} We release TrajectoryBench, a benchmark for long-horizon video generation with complex camera trajectories.

\section{Related Work}
\label{sec:related_works}

\mypar{Post-training for 3D consistency of video models.} 
While some prior works improve 3D consistency via architecture ~\citep{xu2024camco}, memory~\citep{ren2025gen3c}, or training objective~\citep{xiang2026videoweave}, a large number of works perform post-training.
They propose geometry-based reward models, such as epipolar constraint in Epipolar-DPO~\citep{kupyn2025epipolar}, depth reprojection in VGGRPO~\citep{an2026vggrpo}, VideoGPA~\citep{du2026videogpa}, VIGOR~\citep{yin2026vigor} and GeoVideo~\citep{bai2026geovideo}, or Gaussian splatting~\citep{kerbl20233d} in World-R1\citep{wang2026world}. Methods like CamVerse~\citep{wang2026taming}, CamPilot~\citep{ge2026campilot} use similar rewards to post-train for camera control.
While prior works show the efficacy of geometry rewards, they focus on the short-horizon, 80-frame WAN model~\citep{wan2025wan}, and are limited to simple camera trajectories.

\mypar{Credit assignment in RL.}
Credit assignment in RL has proven impactful in LLM reasoning \citep{uesato2022solving, tao2026trace}. In computer vision, reward localization has been explored for certain tasks, as in SpatialFlow-GRPO~\citep{yang2026spatialflow} for editing, LongCat ~\citep{team2026longcat} for omni model post-training, and CreFlow~\citep{ni2026creflow} for robotic video generation.
While they adopt some version of localized reward like loss masking~\citep{ni2026creflow}, they do not comprehensively explore credit assignment.
In this work, we focus on a reward design that incorporates proper credit assignment, which is central to improving 3D consistency over long horizons.

\mypar{Long-horizon video generation post-training.}
Recent efforts enable long-horizon generation via autoregressive rollout of flow-matching chunks~\citep{shen2026lyra2, sun2025worldplay, lingbot-world-v2}. While these models have been post-trained, existing methods are limited to specific aspects. \citep{wang2026worldcompass} and DreamX-World 1.0~\citep{team2026dreamx} post-train only for camera following, a ``local'' property that can be evaluated per-frame without considering the full video. \cite{wang2026worldcompass} train on individual sampled autoregressive chunks, without considering credit assignment across the video. \cite{gu2026worldcycle} use post-training to make reversible actions ``cancel'' each other over long horizons, which differ from our focus on 3D consistency and credit assignment.
\section{Method}
\label{sec:method}

The multifaceted challenge of improving 3D consistency in long-horizon video generation motivates each component of the \method recipe. We explain each component below: \textbf{(1)} spatially-localized credit assignment in \S\ref{sec:local}; \textbf{(2)} local-global mixing in \S\ref{sec:logoblending}; \textbf{(3)} reward interleaving in \S\ref{sec:interleaving}.

\subsection{Reward localization in 3D}
\label{sec:local}

\begin{figure}[t]
\includegraphics[width=\textwidth]{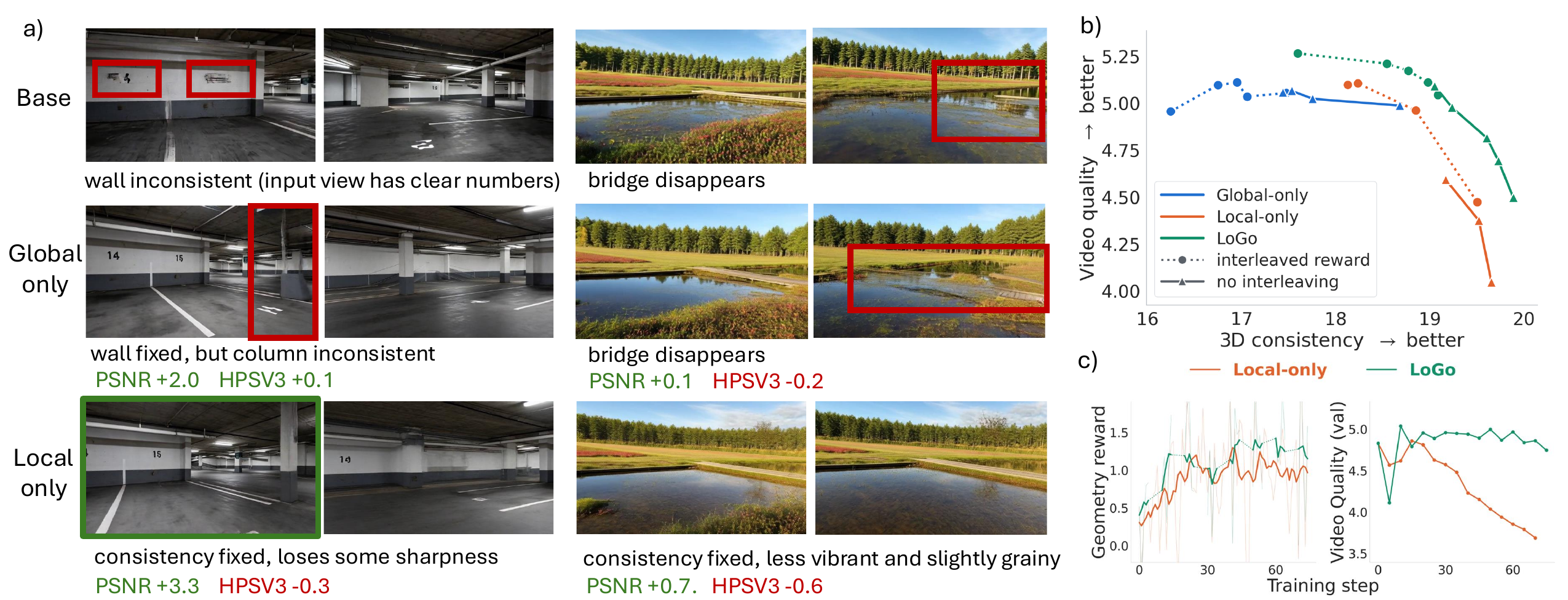}
\vspace{-6mm}
\caption{Motivation of \method. \textbf{(a)} A global reward fails to improve 3D consistency; local reward improves consistency, measured in PSNR, but degrades video quality, measured in HPSv3. This motivates combining the two in \method. \textbf{(b)} \method (green) extends the Pareto frontier in the consistency-quality tradeoff. \textbf{(c)} Reward curves and quality evaluation during training: local reward degrades video quality over time; \method improves consistency without compromising quality.}
\label{fig:method_motivation}
\vspace{-6mm}
\end{figure}

Given a base model and a 3D consistency objective, the naive approach is to use the objective as a scalar reward for post-training. However, as shown in Fig.~\ref{fig:motivation}, such a reward focuses on the overall scene structure, ``washing out'' local inconsistencies. As a result, training only with a global reward is ineffective: as shown in Fig.~\ref{fig:method_motivation}, a global reward fails to fix existing errors (in the lake example, the bridge still disappears) or introduces new ones (in the garage example, it morphs the column).

%% -------- FIGURE -------- %%
%% ---------------- %%

Modern video models fail in three, largely local, ways: \textbf{(1)} inconsistent object appearance, \textbf{(2)} hallucinated or disappearing objects, and \textbf{(3)} local artifacts. A good reward should penalize these failures directly rather than merely ranking full-length videos. Motivated by this, we design a reward localization method that operates directly in 3D space for accurate credit assignment during post-training.

\begin{wrapfigure}{r}{0.6\textwidth}
  \vspace{-7mm}
  \centering
  \includegraphics[width=0.53\textwidth]{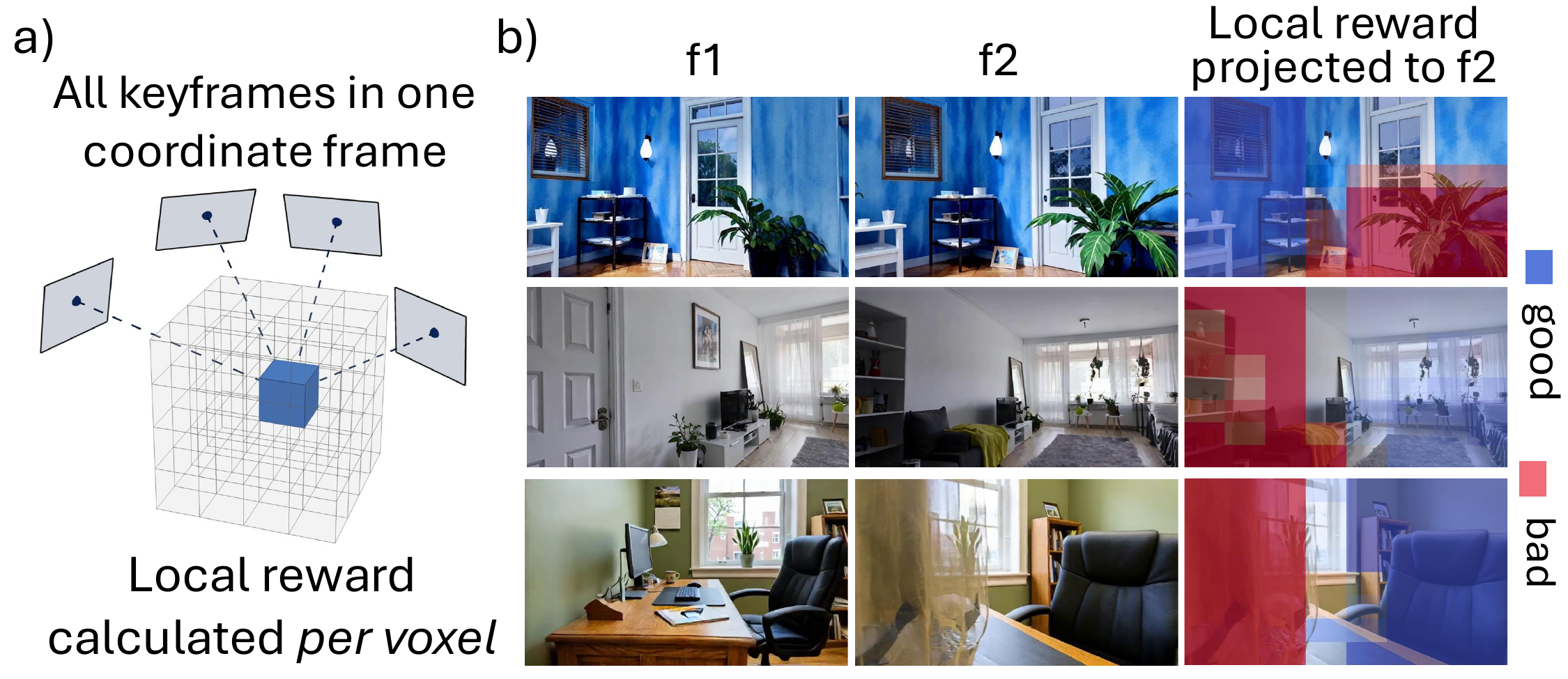}
\vspace{-3mm}
\caption{\textbf{a)}: illustration of voxel reward aggregation. \textbf{b)}: visualization of local reward projected to video frames.}
\vspace{-6mm}
\label{fig:reward_viz}
\end{wrapfigure}

Given a video, we first construct a scene point cloud by passing keyframes to VGGT-$\Omega$~\citep{wang2025vggt} and unproject each pixel using the predicted depth into a shared coordinate system. As shown in Fig.~\ref{fig:reward_viz}a, we voxelize this shared 3D space into a voxel grid, and calculate a spatially localized error per voxel: each voxel's error is defined as the average reprojection error of every pixel unprojected into the voxel. For a pixel, the reprojection error is defined as a weighted combination of RGB error, \ie RGB projection of the scene point cloud into that pixel compared with the pixel's generated RGB, and depth error, \ie depth projection of the scene point cloud into that pixel compared to the pixel's VGGT-predicted depth.

Let $\mathcal{P}$ be the scene point cloud, $c_j$ the camera pose of frame $j$, and $I_j(x)$ the pixel of the $j$-th frame at location $x$. For each voxel $i$, $\mathcal{S}_i$ denotes the set of pixels that unproject into the voxel. If $\text{Proj}_\text{RGB}$ and $\text{Proj}_\text{D}$ is the RGB and depth projection function, respectively, the local voxel error is defined as:
\begin{align}
e_i=\frac{1}{\vert \mathcal{S}_i\vert}\sum_{(x, j) \in \mathcal{S}_i}\alpha \left\lVert \text{Proj}_{\text{RGB}}[\mathcal{P}, c_j, x]-I_{j}^\text{RGB}{(x)}\right\rVert_2^2 + (1-\alpha) \left\vert \text{Proj}_\text{D} [\mathcal{P}, c_j, x] - I_j^\text{D}(x) \right\vert
\end{align}
For small group sizes during post-training, a cross-voxel normalization per rollout can be added to the score to highlight the difference between systematically good or bad rollouts. In our experiments, we set the voxel size to $0.1\times$ the 90-percentile (P90) scene depth. For a 240-frame video, this yields a median of roughly 3,000 voxels per scene, with a median of 200 points per voxel.

\setlength{\columnsep}{6pt} % removes white space to the left

We visualize our local reward via projected heatmaps in Figure~\ref{fig:reward_viz}, with voxel advantages projected onto keyframe patches. Our reward localization scheme can correctly capture each category of local inconsistency that we observe, including appearance inconsistency of the plant in the first example, hallucinated objects in the second example of the suddenly-appearing bookshelf, as well as local artifacts, as seen in the last example as a yellow blob on the left side of keyframe 2.

\subsection{\method: Blending local and global reward}
\label{sec:logoblending}

While the local reward is highly effective at improving 3D consistency, using it in isolation causes video quality to degrade over the course of training (Fig.~\ref{fig:method_motivation}c). We attribute this to reward hacking: as in LLMs, where dense rewards can induce such behavior~\citep{zhuang2026workforceagent}, the local reward fixes inconsistencies at the expense of overall quality, reducing sharpness and vibrancy (Fig.~\ref{fig:method_motivation}a). The pattern is consistent with reward hacking, since blurrier, smoother edges lower depth errors and less vibrant (closer-to-grey) colors lower RGB errors.

In contrast, a global reward, defined as the average reprojection error across all pixels and frames, proves more robust to this degradation. The Pareto curves in Fig.~\ref{fig:method_motivation}b make the trade-off clear: the local reward (orange) sacrifices video quality (measured by HPSv3~\citep{ma2025hpsv3}) for geometric consistency (measured by reprojection PSNR), whereas the global reward (blue) largely preserves quality. To capture both the geometric consistency gains of the local reward and the video quality of the global reward, we blend them as a weighted sum of the normalized reward -- the optimality probability in DiffusionNFT~\citep{zheng2026diffusionnft}.
\begin{align}
    r_{i, \text{LoGo}}=\tfrac{1}{2} + \tfrac{1}{4}\,\mathrm{clip}\!\left(\frac{\bar e_{i,\text{local}} - e_{i,\text{local}}}{Z_{e_{i,\text{local}}}},\,-1,\,1\right) + \tfrac{1}{4}\,\mathrm{clip}\!\left(\frac{\bar e_{\text{global}} - e_{\text{global}}}{Z_{e_{\text{global}}}},\,-1,\,1\right)
\end{align}

This blended local-global reward expands the Pareto frontier in the consistency-quality tradeoff, shown with green in Fig.~\ref{fig:method_motivation}b. Over the course of training, \method improves the geometry reward, \ie the weighted sum of global reward and average local reward across voxels, without degrading aesthetics, as show in Fig.~\ref{fig:method_motivation}c.

\subsection{Reward interleaving}
\label{sec:interleaving}

To ensure our post-trained models are robust across multiple axes, such as video quality and camera control, we interleave post-training with rewards targeting these properties. We adopt a cyclic schedule: $k$ steps with $r_\text{\method}$, then $m$ steps of aesthetic reward -- HPSv3~\citep{ma2025hpsv3} -- and $n$ steps of camera reward -- equally weighted sum of rotation and translation errors, which preserves or improves other metrics while still improving consistency. 
In Fig.~\ref{fig:method_motivation}b, the dotted half of each curve uses reward interleaving, and the solid half does not. Our main results use DiffusionNFT~\citep{zheng2026diffusionnft}; in the Appendix we further demonstrate efficacy with Flow-GRPO~\citep{liu2026flow} and DRaFT~\citep{clark2024directly}.

\mypar{Efficiency. }
\method adds minimal overhead -- only $2.1\%$ of the total reward calculation time and $0.5\%$ of the total per-step training time. \method's overhead comes only from computing the pixel-voxel correspondence (9 seconds per step out of 1920 seconds of total per-step training time on 64 H100s). Training time is dominated by the video rollout, VGGT invocation and backpropagation, all independent of \method.
\section{Experiments}

We now turn to experiments to demonstrate the efficacy of \method. We first introduce TrajectoryBench, which addresses the evaluation gap for long-horizon video generation and complex camera control. Then, we compare \method against post-training baselines across three base video models, showing a consistent advantage. Finally, we ablate \method's design choices, isolating the impact of a spatially localized reward over a global reward on long-horizon generation.

\subsection{Benchmarks}

\mypar{TrajectoryBench.}
We introduce TrajectoryBench, a new benchmark for long-horizon video generation with complex camera trajectories. Compared to existing benchmarks~\citep{duan2025worldscore}, it contains challenging scenarios along three axes: \textbf{(1)} trajectories that move far from the initial view, testing whether a model can generate unseen content; \textbf{(2)} trajectories that revisit the same area from multiple viewing angles, testing 3D consistency; and \textbf{(3)} trajectories with space ``transitions,'' such as passing through a door, testing whether a model can synthesize entirely new scenes.

\setlength{\columnsep}{5pt} % adds white space to the left
\begin{wrapfigure}{r}{0.43\textwidth}
  \vspace{-6mm}
  \centering
  \includegraphics[width=0.4\textwidth]{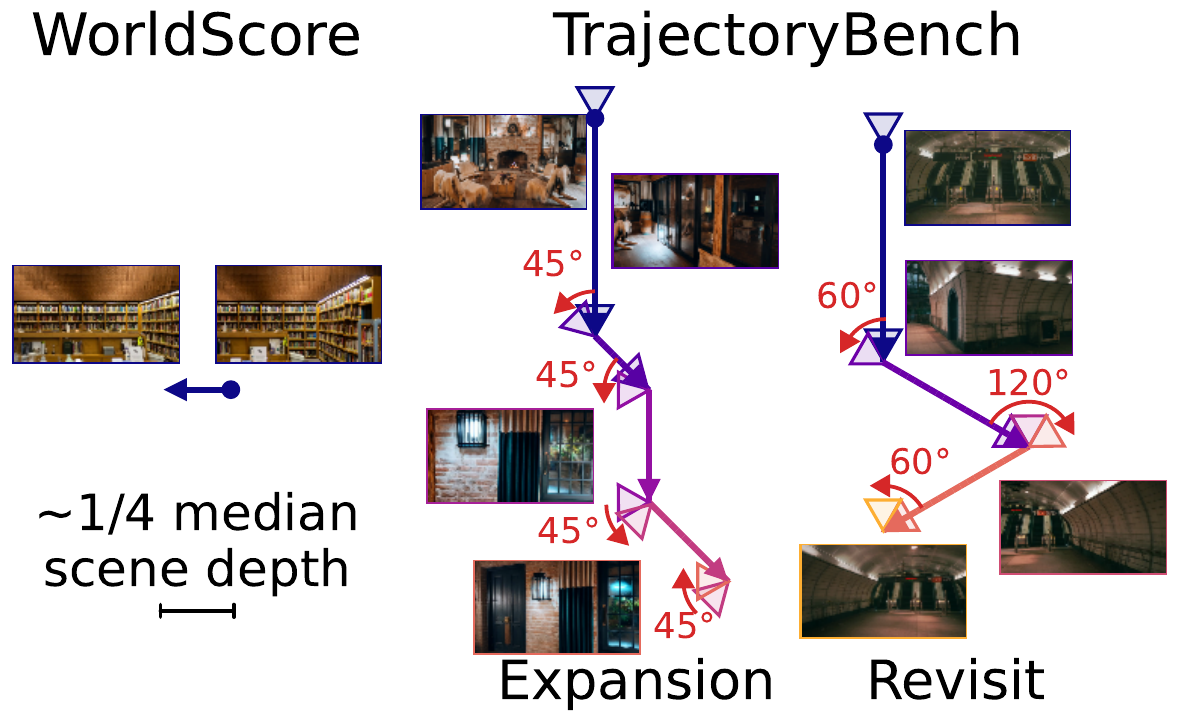}
  \vspace{-3mm}
\caption{Camera trajectories of TrajectoryBench vs. WorldScore.}
\vspace{-3mm}
\label{fig:trajbench}
\end{wrapfigure}
TrajectoryBench contains 2{,}000 examples across three difficulty levels (\emph{easy, medium, hard}) and three space types (indoor, outdoor, transition), each pairing an input image with a camera trajectory. The easy split is subsampled from WorldScore (80 frames, one camera motion) for compatibility with existing benchmarks; medium includes transition scenes that go through a door into a new space, and contains concatenated three single-camera-motion commands from WorldScore for indoor and outdoor scens; and hard split packs seven to eight camera motions into 240 frames, using our own sourced initial images. Fig.~\ref{fig:trajbench} contrasts a WorldScore trajectory (easy) with a hard TrajectoryBench one in expansion and revisit modes: WorldScore trajectories are simple, whereas TrajectoryBench tests generation from views far from the initial frame or repeated revisits. TrajectoryBench also provides comprehensive metrics spanning 3D consistency, camera-following, and video quality.

\mypar{DL3DV.}
We evaluate on the DL3DV~\citep{ling2024dl3dv} 100-scene holdout set used by prior works~\citep{du2026videogpa}. However,~\cite{du2026videogpa} evaluate with only the initial image and text prompt, and omit camera control. Since our focus is on camera-controlled video models, we additionally condition on DL3DV's original camera trajectories and evaluate camera control accuracy.

\subsection{Training and evaluation settings}
\mypar{Base models and training setting.}
We post-train three state-of-the-art camera-controlled video generation base models: (1) Lingbot-world-v2~\citep{lingbot-world-v2}, a 14B autoregressive model that generates 12 frames at a time with KV cache, (2) Lyra-2~\citep{shen2026lyra2}, a 14B model that generates 80 frames per chunk, with a sparse 3D cache that supports camera-based retrieval; and (3) UniWorld~\citep{zhou2026uniworldview}, a 14B bidirectional model trained on 80-frame clips. Because UniWorld's quality degrades greatly beyond 80 frames, we subsample longer trajectories into 80-frame generations.
% Although it is trained on 80-frame videos and perform poorly on longer videos, we include it for completeness since it is the most recent winner of WorldScore~\citep{duan2025worldscore}. We evaluate it by subsampling long trajectories into 80 frames.
We train all models on 1700 DL3DV scenes until convergence, which takes 16-36 hours on 64 H100s. More details can be found in \S\ref{sec:additional_training_details} in the Appendix.

\mypar{Metrics.}
We evaluate 3D consistency, camera following, and video quality.
For \textbf{3D consistency}, we measure RGBD reprojection PSNR~\citep{bai2026geovideo, du2026videogpa}, its depth-only variant MVCS~\citep{bai2026geovideo}, Gaussian reconstruction error~\citep{wang2026worldr1}, and epipolar error~\citep{kupyn2025epipolar}. We report depth metrics using VGGT~\citep{wang2025vggt} and Depth Anything V3~\citep{lin2025depth} to avoid model bias.
For \textbf{camera control}, we report relative pose error (RPE): the VGGT-estimated camera translation and rotation between frames versus the ground truth.
For \textbf{video quality}, we measure video quality (VQ) score from VideoReward~\citep{liu2026improving}, a Qwen-based~\citep{yang2024qwen2technicalreport} reward model trained on human preference.

\mypar{Baselines.}
We compare with state-of-the-art 3D consistency post-training methods. Prior methods are based on WAN~\citep{wan2025wan}, a 80-frame image/text-to-video model without camera control. Since there has been no prior work on post-training camera-controlled models for 3D consistency, we extend these methods to this setting. We compare to the following: \textbf{(1)} VideoGPA~\citep{du2026videogpa}, a DPO~\citep{rafailov2023direct} method using a single scalar VGGT~\citep{wang2025vggt} reprojection reward for a whole video;
\textbf{(2)} World-R1~\citep{zhou2026uniworldview} performs Flow-GRPO~\citep{liu2026flow} using VLM feedback on Gaussian reconstruction, yielding a scalar reward per rollout.
%While VideoGPA is purely depth-based and World-R1 Gaussian-based, we evaluate both depth reprojection and Gaussian reconstruction errors to be comprehensive.

\subsection{Results}

\begin{table}[t!]
\centering
\small
\setlength{\tabcolsep}{5pt}
\newcommand{\std}[1]{{\tiny$\pm$#1}}
\resizebox{\linewidth}{!}{
\begin{tabular}{@{}ll *{8}{c}@{}}
\toprule
& \multirow{2}{*}{Model} & \multicolumn{5}{c}{3D consistency} & \multicolumn{2}{c}{Camera}
 & Video quality \\
 \cmidrule(lr){3-7} \cmidrule(lr){8-9} \cmidrule(lr){10-10}
 & & PSNR-V $\uparrow$ & PSNR-D $\uparrow$ & PSNR-GS $\uparrow$ & MVCS $\uparrow$ & Epipolar $\downarrow$ & RPE$^\circ$ $\downarrow$ & RPE trans $\downarrow$ & VR-VQ $\uparrow$ \\
\midrule
\multirow{5}{*}{Lingbot2} & Base & 16.9 & 16.0 & 16.9 & 0.842 & 1.507 & 13.2 & 0.108 & 0.21\\
 & VideoGPA & 17.1 & 16.1 & 17.0 & 0.847 & 1.486 & 13.1 & 0.107 & 0.22\\
 & World-R1 & 17.0 & 16.0 & 16.9 & 0.839 & 1.515 & 13.1 & 0.108 & 0.21\\
 & Global-only (ours) & 18.0 & 17.1 & 17.8 & 0.872 & 1.306 & 11.2 & 0.102 & 0.31\\
 & \method (ours) & \textbf{19.0} & \textbf{18.2} & \textbf{18.4} & \textbf{0.894} & \textbf{1.111} & 8.9 & 0.103 & 0.37\\
\cmidrule(lr){1-10}
\multirow{5}{*}{UniWorld} & Base & 17.6 & 16.8 & 17.5 & 0.881 & 1.260 & 13.8 & 0.132 & 0.39\\
 & VideoGPA & 17.7 & 16.9 & 17.5 & 0.882 & 1.275 & 13.8 & 0.132 & 0.39\\
 & World-R1 & 17.9 & 17.1 & 18.2 & 0.869 & 1.401 & 13.1 & 0.129 & 0.22\\
 & Global-only (ours) & 18.3 & 17.5 & 18.0 & 0.892 & 1.195 & 13.3 & 0.126 & 0.37\\
 & \method (ours)& \textbf{18.8} & \textbf{18.0} & \textbf{18.6} & \textbf{0.901} & \textbf{1.108} & 14.0 & 0.130 & 0.41\\
\cmidrule(lr){1-10}
\multirow{5}{*}{Lyra2} & Base & 18.4 & 17.7 & 19.0 & 0.897 & 1.550 & 7.9 & 0.100 & -0.23\\
 & VideoGPA & 18.0 & 17.4 & 18.6 & 0.896 & 1.511 & 7.7 & 0.099 & -0.21\\
 & World-R1 & 18.8 & 18.1 & 19.4 & 0.899 & 1.550 & 7.7 & 0.101 & -0.20\\
 & Global-only (ours) & 19.4 & 18.6 & 19.7 & 0.900 & 1.396 & 7.6 & 0.099 & -0.21\\
 & \method (ours) & \textbf{19.7} & \textbf{19.0} & \textbf{20.0} & \textbf{0.905} & \textbf{1.352} & 7.5 & 0.101 & -0.19\\
\bottomrule
\end{tabular}
}
\vspace{-3mm}
\caption{\textbf{Results on TrajectoryBench.} PSNR-V and PSNR-D measure RGBD reprojection PSNR (using VGGT and DA3), PSNR-GS is based on Gaussian reconstruction. \method achieves strong gains in all consistency metrics across three base models. It increases PSNR-D by 2.2dB from Lingbot2 base model, and reduces epipolar error by 26\%. Meanwhile, \method preserves, and often improves, camera following and video quality. Standard errors are in Table~\ref{tab:trajbenchfull} in the Appendix.}
\vspace{-2mm}
\label{tab:trajbench_merge}
\end{table}

\begin{table*}[t!]
    \centering
    \small
    \setlength{\tabcolsep}{4pt}
    \newcommand{\std}[1]{{\tiny$\pm$#1}}
    \resizebox{\linewidth}{!}{
     \begin{tabular}{@{}ll *{8}{c}@{}}
    \toprule
    & \multirow{2}{*}{Model}
     & \multicolumn{5}{c}{3D consistency}
      & \multicolumn{2}{c}{Camera}
      & \multicolumn{1}{c}{Video quality} \\
     \cmidrule(lr){3-7} \cmidrule(lr){8-9} \cmidrule(lr){10-10}
     & & PSNR-V $\uparrow$ & PSNR-D $\uparrow$ & PSNR-GS $\uparrow$ & MVCS $\uparrow$ & Epipolar $\downarrow$ & RPE$^\circ$ $\downarrow$ & RPE trans $\downarrow$ & VR-VQ $\uparrow$ \\
    \midrule
    \multirow{5}{*}{Lingbot2} & Base & 13.3 & 13.1 & 15.6 & 0.719 & 2.161 & 20.52 & 0.082 & -0.422\\
 & VideoGPA & 13.3 & 13.2 & 15.6 & 0.728 & 2.429 & 20.15 & 0.081 & -0.409\\
 & World-R1 & 13.6 & 13.3 & 15.7 & 0.717 & 2.271 & 20.13 & 0.070 & -0.429\\
 & Global-only (ours) & 14.0 & 13.8 & 16.3 & 0.772 & 2.107 & 22.68 & 0.086 & -0.305\\
 & \method (ours) & \textbf{16.1} & \textbf{15.8} & \textbf{17.3} & \textbf{0.795} & 1.365 & 18.33 & 0.077 & -0.283\\
\cmidrule(lr){1-10}
\multirow{5}{*}{UniWorld} & Base & 14.7 & 14.5 & 16.6 & 0.748 & 1.510 & 12.94 & 0.057 & -0.128\\
 & VideoGPA & 14.7 & 14.6 & 16.7 & 0.752 & 1.623 & 13.31 & 0.058 & -0.129\\
 & World-R1 & 14.9 & 14.7 & 17.3 & 0.747 & 1.926 & 15.45 & 0.064 & -0.234\\
 & Global-only (ours) & 14.9 & 14.7 & 16.7 & 0.778 & 1.557 & 12.18 & 0.058 & -0.134\\
 & \method (ours) & \textbf{15.4} & \textbf{15.1} & \textbf{17.4} & \textbf{0.807} & 1.429 & 12.64 & 0.055 & -0.121\\
\cmidrule(lr){1-10}
\multirow{5}{*}{Lyra2} & Base & 15.2 & 14.8 & 16.8 & 0.819 & 2.448 & 22.72 & 0.148 & -0.557\\
 & VideoGPA & 15.0 & 14.7 & 16.6 & 0.813 & 2.595 & 21.95 & 0.143 & -0.508\\
 & World-R1 & 15.3 & 15.2 & 16.9 & 0.832 & 2.213 & 21.72 & 0.140 & -0.523\\
 & Global-only (ours) & 15.6 & 15.3 & 17.1 & 0.826 & 2.244 & 22.17 & 0.145 & -0.525\\
 & \method (ours) & \textbf{15.9} & \textbf{15.6} & \textbf{17.3} & \textbf{0.844} & 1.863 & 22.38 & 0.145 & -0.513\\
\bottomrule
    \end{tabular}
    }
\vspace{-3mm}
\caption{\textbf{Results on DL3DV.} \method yields larger gains than baseline methods across three base models for all 3D consistency metrics, while maintaining camera control and video quality, with up to $37\%$ decrease in epipolar error. Standard errors are shown in Table~\ref{tab:dl3dv_se} in the Appendix.}
    \label{tab:dl3dv}
\vspace{-6mm}
\end{table*}

%% -------- FIGURE -------- %%
\begin{figure*}[p]
\centering
\includegraphics[width=\textwidth]{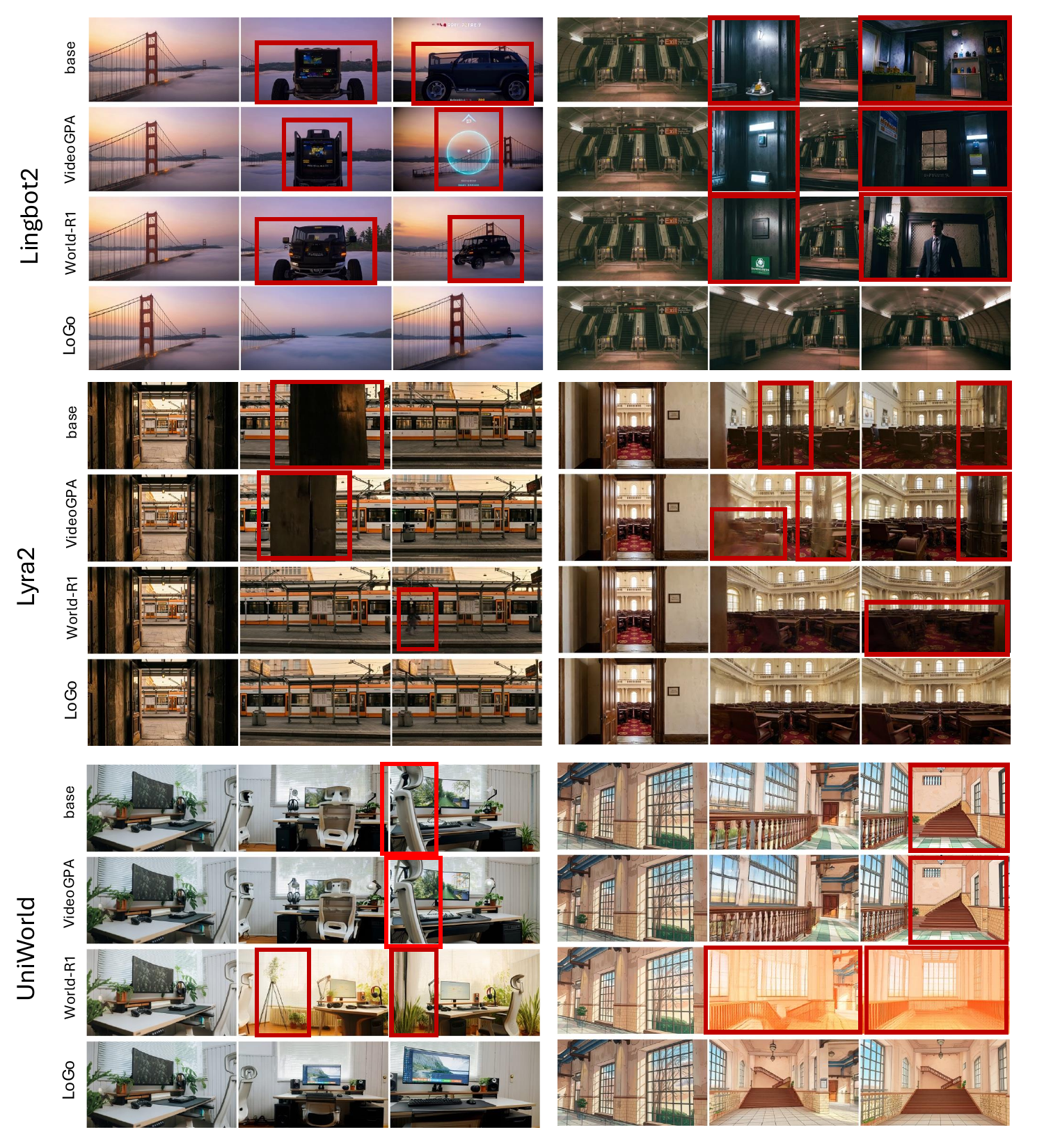}
\vspace{-8mm}
\caption{\textbf{Qualitative results of \method vs. baselines}. Red boxes indicate inconsistent regions. \method effectively fixes both global and local inconsistencies, including scene change (top right), hallucinated objects (top left), floater artifacts (middle), and inconsistent object appearance (bottom). Baseline post-training methods are not capable of resolving these issues, while \method is.}
\label{fig:qualitative}
%% ---------------- %%
\vfill
\centering
\includegraphics[width=\textwidth]{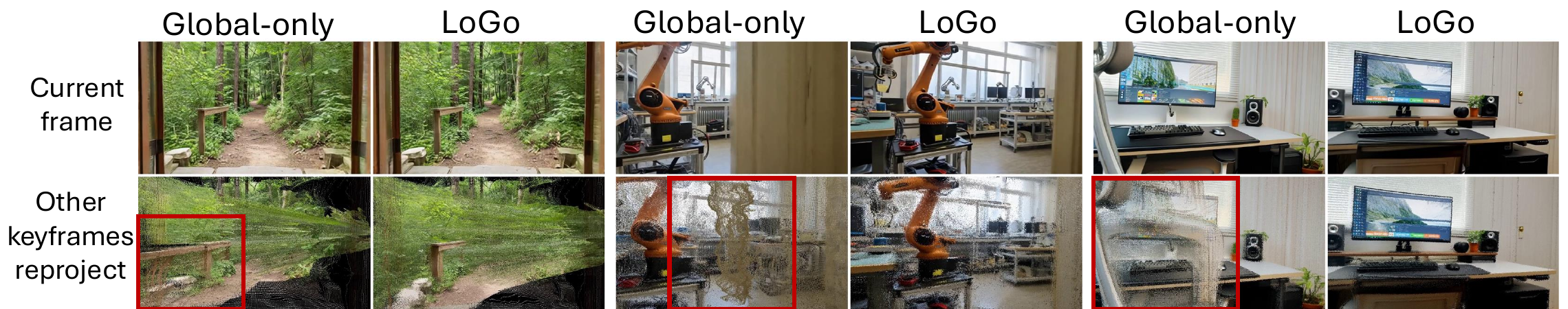}
\caption{\textbf{Global-only reward vs. \method.} A 3D-consistent video should closely match its reprojections, though not perfectly, since keyframes in long videos overlap only sparsely. Global reward incurs inconsistencies, highlighted in red, such as hallucinated object (double railing), artifact (floater in the robot scene), and object appearance change (chair). \method, by combining global reward with local reward, effectively removes these inconsistencies.}
\label{fig:global_logo}
\end{figure*}
%% ---------------- %%

\mypar{\method post-training improves consistency, while maintaining or improving video quality and camera control.}
On both TrajectoryBench (Table~\ref{tab:trajbench_merge}) and DL3DV (Table~\ref{tab:dl3dv}), \method significantly improves 3D consistency from the base models, with up to +2.7dB PSNR and 37\% decrease of epipolar error. For the best model, Lingbot2, \method also improves camera control and video quality.
\method's effectiveness is shown qualitatively in Fig.~\ref{fig:qualitative}, addressing both global and local inconsistencies, including scene change (subway changes into a room, top-right), hallucinated objects (car appears in golden gate bridge scene, top left), artifacts (column-like floaters for Lyra2), and object appearance consistency (chair orientation change, bottom left). All these inconsistencies are fixed by \method. Full videos can be found on our project website.

\mypar{\method shows stronger gains than prior post-training methods, partly due to credit assignment.}
Prior methods, VideoGPA and World-R1, show moderate gains from base models. For Lingbot2, on both benchmarks the prior methods improve by 0.1dB for PSNR-GS and by 0.2dB for PSNR-D. We hypothesize \method's advantage comes partly from credit assignment via the local reward. In Table~\ref{tab:dl3dv}, our global-only variant's advantage over baselines is moderate, whereas \method shows stronger gains. This comparison showcases the importance of credit assignment via the local reward. 

Fig.~\ref{fig:qualitative} shows qualitatively that baseline methods fail to fix the base models' inconsistencies: in the golden gate bridge example, both VideoGPA and World-R1 change the hallucinated object but do not remove it.
World-R1, while reducing floaters (middle left), generates other artifacts like a morphing human-shadow-like objects in the train station and morphs the chairs in the auditorium (middle right). In the cartoon stylized example, UniWorld base model and all baselines transform the hallway with door into a staircase scene, while \method is able to generate a coherent stair scene. 

Fig.~\ref{fig:global_logo} further compares \method with our global-only variant qualitatively, visualizing the point cloud reprojection from other keyframes versus the generated frame. A global-only reward fails to fix local inconsistencies like double railing, artifacts, and the altered chair orientation. This corroborates the quantitative finding that credit assignment is an important contributor to \method's advantage.

\begin{table}[t!]
\centering
\small
\setlength{\tabcolsep}{5pt}
\newcommand{\std}[1]{{\tiny$\pm$#1}}
\resizebox{0.75\linewidth}{!}{
\begin{tabular}{@{}ll *{9}{c}@{}}
\toprule
& \multirow{2}{*}{Model} & \multicolumn{3}{c}{PSNR-V $\uparrow$} & \multicolumn{3}{c}{PSNR-D $\uparrow$} & \multicolumn{3}{c}{Epipolar $\downarrow$} \\
 \cmidrule(lr){3-5} \cmidrule(lr){6-8} \cmidrule(lr){9-11}
 & & E & M & H & E & M & H & E & M & H \\
\midrule
\multirow{4}{*}{Lingbot2} & Base & 22.6 & 18.2 & 15.3 & 20.6 & 17.6 & 14.5 & 0.524 & 0.681 & 2.034\\
 & VideoGPA & 22.7 & 18.3 & 15.4 & 20.7 & 17.7 & 14.6 & 0.533 & 0.684 & 1.997\\
 & World-R1 & 22.7 & 18.1 & 15.4 & 20.6 & 17.4 & 14.5 & 0.526 & 0.681 & 2.046\\
 & \method & \textbf{23.8} & \textbf{20.2} & \textbf{17.6} & \textbf{21.6} & \textbf{19.7} & \textbf{17.0} & \textbf{0.490} & \textbf{0.601} & \textbf{1.439}\\
\midrule
\multirow{4}{*}{UniWorld} & Base & 23.0 & 17.2 & 16.7 & 21.2 & 16.4 & 16.1 & 0.450 & \textbf{0.760} & 1.623\\
 & VideoGPA & 23.1 & 17.2 & 16.8 & 21.2 & 16.4 & 16.2 & 0.450 & 0.773 & 1.642\\
 & World-R1 & 22.4 & 17.3 & 17.2 & 20.5 & 16.5 & 16.6 & 0.462 & 0.860 & 1.805\\
 & \method & \textbf{23.2} & \textbf{18.2} & \textbf{18.1} & \textbf{21.3} & \textbf{17.4} & \textbf{17.5} & \textbf{0.446} & \textbf{0.760} & \textbf{1.379}\\
\midrule
\multirow{4}{*}{Lyra2} & Base & 22.2 & 17.9 & 17.8 & 21.1 & 17.2 & 17.3 & 0.519 & 0.874 & 2.027\\
 & VideoGPA & 22.1 & 17.7 & 17.3 & 21.1 & 17.1 & 16.7 & 0.519 & 0.878 & 1.963\\
 & World-R1 & 22.2 & 18.1 & 18.4 & 21.0 & 17.5 & 17.8 & 0.529 & 0.939 & 1.999\\
 & \method & \textbf{22.5} & \textbf{18.8} & \textbf{19.4} & \textbf{21.2} & \textbf{18.1} & \textbf{18.8} & \textbf{0.510} & \textbf{0.847} & \textbf{1.723}\\
\bottomrule
\end{tabular}
}
\vspace{-2mm}
\caption{\textbf{Difficulty breakdown on TrajectoryBench.} We show results on easy, medium, and hard categories of TrajectoryBench. PSNR-V denotes RGBD reprojection error with VGGT, and PSNR-D with DA3. \method's margin over the baselines widens as difficulty increases across all three base models. This trend also holds for other metrics found in Table~\ref{tab:trajbenchfull} in the Appendix.}
\label{tab:trajbench_tiered}
\vspace{-18pt}
\end{table}

\mypar{\method's advantage widens as the trajectories become more difficult.}
TrajectoryBench's increasing difficulty tier allows us to further understand \method's advantage. While all metrics degrade as the difficulty increases from easy to difficult, as expected, Table~\ref{tab:trajbench_tiered} shows that \method's advantage grows with difficulty. While \method's advantage is moderate on easy trajectories (+1dB PSNR-D for Lingbot2), the gain is much higher on hard trajectories (+2.5dB).
The easy/medium/hard breakdown for all metrics with standard errors are shown in Table~\ref{tab:trajbenchfull} in the Appendix.

\section{Ablations and Analyses}

\setlength{\columnsep}{7pt} % removes white space to the left
\begin{wrapfigure}{r}{0.35\textwidth}
  \vspace{-0mm}
  \centering
 \includegraphics[width=0.3\textwidth]{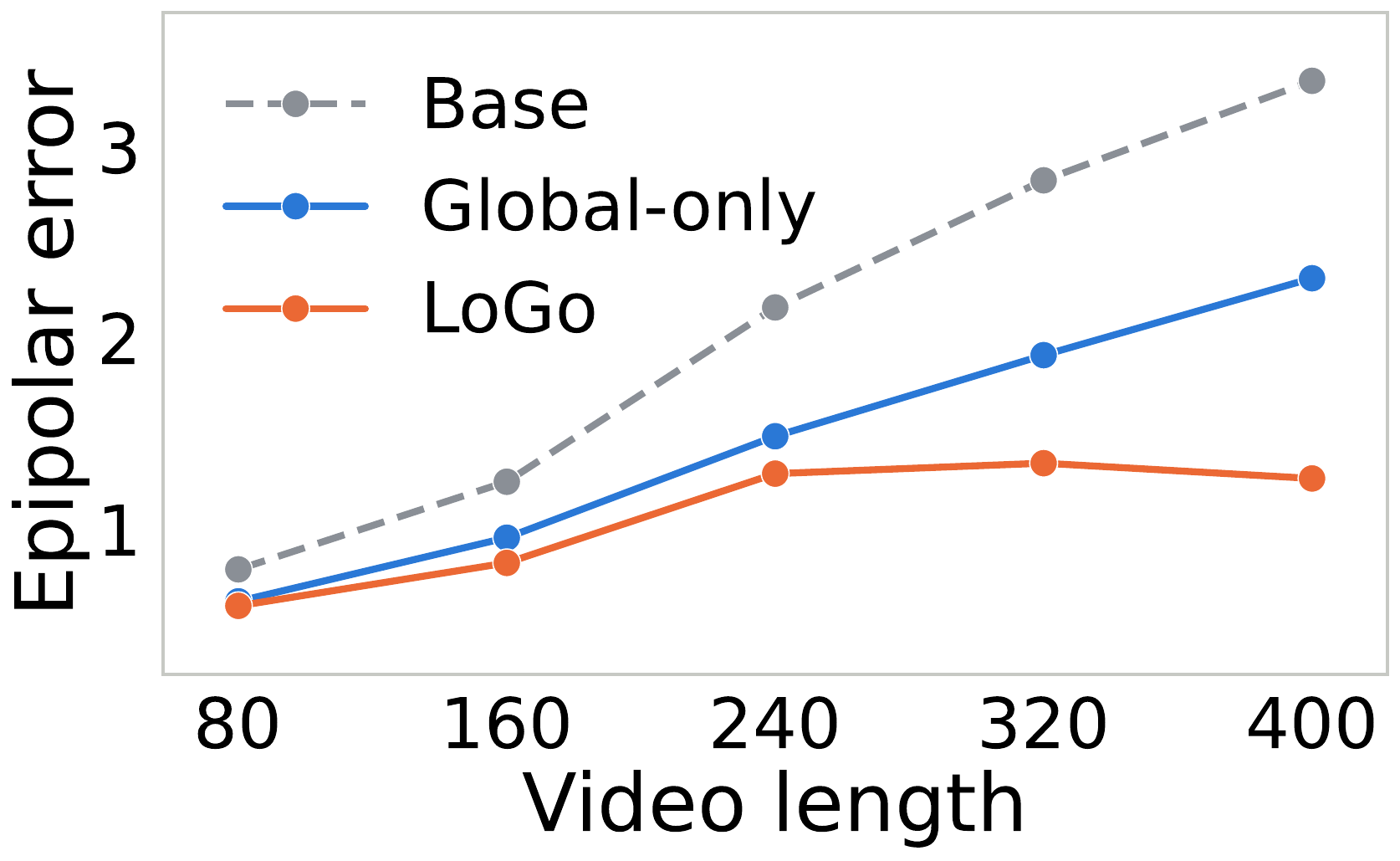}
   \vspace{-4mm}
\caption{Epipolar error vs. length.}
\label{fig:vary_vid_length}
  \vspace{-6mm}
\end{wrapfigure}

We analyze the performance of \method over increasing video lengths, compare different credit assignment methods for \method's local reward, and ablate each component of \method's recipe. We further provide an ablation of voxel granularity and show \method's robustness on other RL algorithms.

\mypar{\method's gain strengthens as video length increases.}
We evaluate videos ranging from 80 to 400 frames in length. As shown in Fig.~\ref{fig:vary_vid_length}, longer-horizon generation is more challenging for the base model, Lingbot2, yielding higher epipolar errors. \method leads at all lengths, and its advantage widens with the growing frame length -- from $24\%$ epipolar error reduction at 80 frames to $62\%$ at 400 frames.

\begin{table}[t!]
\centering
\small
\setlength{\tabcolsep}{4pt}
\newcommand{\std}[1]{{\tiny$\pm$#1}}
\resizebox{\linewidth}{!}{
\begin{tabular}{@{}l *{8}{c}@{}}
\toprule
\multirow{2}{*}{} & \multicolumn{5}{c}{3D Consistency}
  & \multicolumn{2}{c}{Camera} & \multicolumn{1}{c}{Video Quality}\\
 \cmidrule(lr){2-6} \cmidrule(lr){7-8} \cmidrule(lr){9-9}
& PSNR-V $\uparrow$ & PSNR-D $\uparrow$ & PSNR-GS $\uparrow$ & MVCS $\uparrow$ & Epipolar $\downarrow$
  & RPE$^\circ$ $\downarrow$ & RPE trans $\downarrow$ & VR-VQ $\uparrow$\\
\midrule
Base & 15.3 & 14.5 & 15.7 & 0.812 & 2.034 & 13.2 & 0.108 & 0.21\\
Global-only & 16.5 & 15.7 & 16.7 & 0.848 & 1.726 & 11.2 & 0.102 & 0.31\\
Window & 17.0 & 16.3 & 17.1 & 0.856 & 1.666 & 9.5 & \textbf{0.094} & 0.32\\
Framewise & 17.0 & 16.4 & 16.9 & 0.857 & 1.531 & 10.7 & 0.101 & 0.24\\
Patchwise & 16.9 & 16.2 & 17.0 & 0.865 & 1.660 & \textbf{8.4} & 0.100 & 0.24\\
Voxel & \textbf{17.8} & \textbf{17.2} & \textbf{17.7} & \textbf{0.874} & \textbf{1.381} & 8.9 & 0.107 & \textbf{0.38}\\
\bottomrule
\end{tabular}
}
\vspace{-3mm}
\caption{\textbf{Comparison of different local rewards.} We compare voxel (used in \method) with framewise, 2D patchwise, and 40-frame window credit assignment. Voxel-based yields the best consistency gains. Easy/medium/hard breakdown and standard errors are in Table~\ref{tab:ablation_ca_full} in the Appendix.}
\label{tab:ablation_ca}
\vspace{-9pt}
\end{table}

\mypar{Voxel credit assignment in 3D wins over frame-or-patch-level assignment.}
There are multiple methods of credit assignment, such as  assiging credit to windows of frames, each frame, or 2D patches, in addition to 3D voxels. Table~\ref{tab:ablation_ca} compares these credit assignment methods, and shows voxels to be the strongest. This is intuitive since voxels occupy the 3D space where inconsistencies occur. Localizing per-frame or per-patch can be inaccurate: since camera following is imperfect, a fixed frame or patch usually does not correspond to the same region across various rollouts.

\begin{table}[t!]
\centering
\small
\setlength{\tabcolsep}{4pt}
\newcommand{\std}[1]{{\tiny$\pm$#1}}
\resizebox{\linewidth}{!}{
\begin{tabular}{@{}l *{9}{c}@{}}
\toprule
\multirow{2}{*}{} & \multicolumn{5}{c}{3D Consistency}
 & \multicolumn{2}{c}{Camera} & \multicolumn{2}{c}{Video Quality}\\
 \cmidrule(lr){2-6} \cmidrule(lr){7-8} \cmidrule(lr){9-10}
 & PSNR-V $\uparrow$ & PSNR-D $\uparrow$ & PSNR-GS $\uparrow$ & MVCS $\uparrow$ & Epipolar $\downarrow$ & RPE$^\circ$ $\downarrow$ & RPE trans $\downarrow$ & VR-VQ $\uparrow$ & HPSv3 $\uparrow$\\
\midrule
Base & 16.9 & 16.0 & 16.9 & 0.842 & 1.507 & 13.2 & 0.108 & 0.21 & 5.09\\
Local-only & \textbf{19.2} & 18.2 & \textbf{18.7} & 0.893 & 1.078 & 11.2 & 0.116 & 0.32 & 4.47\\
+ interleave & \textbf{19.2} & \textbf{18.3} & 18.6 & \textbf{0.897} & \textbf{1.072} & \textbf{8.9} & 0.107 & \textbf{0.38} & 4.94\\
+ global (final) & 19.0 & 18.2 & 18.4 & 0.894 & 1.111 & \textbf{8.9} & \textbf{0.103} & 0.37 & \textbf{5.18}\\
\bottomrule
\end{tabular}
}
\vspace{-3mm}
\caption{\textbf{Ablating the \method recipe.} Using local-only reward yields best 3D consistency but drops camera and aesthetics metrics. Adding in reward interleaving improves other metrics, but still degrades HPSv3 compared to the base model. \method, which includes both reward interleaving and local-global blending, yields the best overall result. Standard errors are in Table~\ref{tab:ablation_recipe_se} in the Appendix.}
\vspace{-6mm}
\label{tab:ablation_recipe}
\end{table}

\mypar{Ablating the \method recipe.}
As explained in \S\ref{sec:logoblending} and \S\ref{sec:interleaving}, local-global blending and reward interleaving improve video quality and camera following. We ablate the effect of each by adding them in one at a time in Table~\ref{tab:ablation_recipe}.
Only using the local voxel reward yields the highest 3D consistency metrics at the cost of camera and aesthetics. Applying reward interleaving without local-global blending improves camera and video quality, but still degrades HPSv3. Blending with global reward further improves camera and aesthetics, surpassing the base model in every metric.

\mypar{\method's advantage persists with other post-training algorithms.}
While our main results use DiffusionNFT~\citep{zheng2026diffusionnft}, we also show \method's advantage with other post-training methods, including FlowGRPO~\citep{liu2026flow} and DRaFT~\citep{clark2024directly} in Table~\ref{tab:ablation_rl_full} in the Appendix.

\mypar{Voxel granularity. }
Across voxel sizes from 0.05$\times$ to 0.7$\times$ 90-percentile (P90) scene depth, finer granularity up to 0.1$\times$ P90 scene depth improves both 3D consistency and camera control (\S\ref{sec:additional_voxel}).
\section{Conclusion and Limitations}

We propose \method, a reward design which enables camera-controlled video models to generate 3D-consistent worlds by post-training with a local-global reward blending. \method, via effective 3D credit assignment, outperforms prior methods. Motivated by the lack of 3D consistency evaluation for long-context camera-controlled video generation, we propose TrajectoryBench, which we hope will serve as a testbed for future development. We release code, checkpoints, and benchmark. 

\method has two main limitations: \textbf{(1)} Very long-horizon generation (beyond 400 frames) remains challenging and might require new memory design, not just better reward design; \textbf{(2)} \method focuses on static scenes, and extending it to dynamic scenes is an exciting direction for future work. There, 4D reconstruction techniques such as~\cite{dart_2026_CVPR} could be used to compute rewards similar in spirit to \method.

\section{AI use statement}
In this paper, we used LLMs to polish language (asking ChatGPT to help with awkward sentences) and for visualization of schematic illustrations. We also used AI coding tools to help speed up standard code implementation. We have not used generative AI tools for the core research idea or to write the content of the paper.
\section{Acknowledgments}
We thank the World Labs team for discussions, questions and comments on the project.

\bibliography{iclr2027_conference}

@article{kupyn2025epipolar,
  title={Epipolar geometry improves video generation models},
  author={Kupyn, Orest and Uscidda, Th{\'e}o and Gazulla, Marta Tintore and Manhardt, Fabian and Tombari, Federico and Rupprecht, Christian},
  journal={arXiv preprint arXiv:2510.21615},
  year={2025}
}

@article{an2026vggrpo,
  title={Vggrpo: Towards world-consistent video generation with 4d latent reward},
  author={An, Zhaochong and Kupyn, Orest and Uscidda, Th{\'e}o and Colaco, Andrea and Ahuja, Karan and Belongie, Serge and Gonzalez-Franco, Mar and Gazulla, Marta Tintore},
  journal={arXiv preprint arXiv:2603.26599},
  year={2026}
}

@article{du2026videogpa,
  title={Videogpa: Distilling geometry priors for 3d-consistent video generation},
  author={Du, Hongyang and Ye, Junjie and Cong, Xiaoyan and Li, Runhao and Ni, Jingcheng and Agarwal, Aman and Zhou, Zeqi and Li, Zekun and Balestriero, Randall and Wang, Yue},
  journal={arXiv preprint arXiv:2601.23286},
  year={2026}
}

@article{yin2026vigor,
  title={VIGOR: VIdeo Geometry-Oriented Reward for Temporal Generative Alignment},
  author={Yin, Tengjiao and Shi, Jinglei and Guo, Heng and Wang, Xi},
  journal={arXiv preprint arXiv:2603.16271},
  year={2026}
}

@article{wang2026world,
  title={World-r1: Reinforcing 3d constraints for text-to-video generation},
  author={Wang, Weijie and He, Xiaoxuan and Gu, Youping and Yang, Yifan and Zhang, Zeyu and He, Yefei and Ding, Yanbo and Hu, Xirui and Chen, Donny Y and He, Zhiyuan and others},
  journal={arXiv preprint arXiv:2604.24764},
  year={2026}
}

@article{uesato2022solving,
  title={Solving math word problems with process-and outcome-based feedback},
  author={Uesato, Jonathan and Kushman, Nate and Kumar, Ramana and Song, Francis and Siegel, Noah and Wang, Lisa and Creswell, Antonia and Irving, Geoffrey and Higgins, Irina},
  journal={arXiv preprint arXiv:2211.14275},
  year={2022}
}

@article{wan2025wan,
  title={Wan: Open and advanced large-scale video generative models},
  author={Wan, Team and Wang, Ang and Ai, Baole and Wen, Bin and Mao, Chaojie and Xie, Chen-Wei and Chen, Di and Yu, Feiwu and Zhao, Haiming and Yang, Jianxiao and others},
  journal={arXiv preprint arXiv:2503.20314},
  year={2025}
}

@article{tao2026trace,
  title={TRACE: Turn-level Reward Assignment via Credit Estimation for Long-Horizon Agents},
  author={Tao, Leitian and Peng, Baolin and Yao, Wenlin and Ge, Tao and Cheng, Hao and Wang, Mike Hang and Gao, Jianfeng and Li, Sharon},
  journal={arXiv preprint arXiv:2607.13988},
  year={2026}
}

@article{yang2026spatialflow,
  title={SpatialFlow-GRPO: Where Spatial Credit Drives Image Editing},
  author={Yang, Yankai and Long, Yancheng and Chen, Wei and Lu, Xingyu and Wei, Hongyang and Wen, Bin and Yang, Fan and Gao, Tingting and Li, Han and Yang, Shuo},
  journal={arXiv preprint arXiv:2606.26872},
  year={2026}
}

@article{ni2026creflow,
  title={CreFlow: Corrective Reflow for Sparse-Reward Embodied Video Diffusion RL},
  author={Ni, Zhenyang and Li, Yijiang and Jiao, Ruochen and Zhan, Simon Sinong and Chen, Sipeng and Yin, Zhenfei and Chen, Minshuo and Torr, Philip and Wang, Zhaoran and Zhu, Qi},
  journal={arXiv preprint arXiv:2605.14274},
  year={2026}
}

@article{wang2026worldcompass,
  title={Worldcompass: Reinforcement learning for long-horizon world models},
  author={Wang, Zehan and Wang, Tengfei and Zhang, Haiyu and Zuo, Xuhui and Wu, Junta and Wang, Haoyuan and Sun, Wenqiang and Wang, Zhenwei and Cao, Chenjie and Zhao, Hengshuang and others},
  journal={arXiv preprint arXiv:2602.09022},
  year={2026}
}

@article{team2026longcat,
  title={LongCat-Video-Avatar 1.5 Technical Report},
  author={Team, Meituan LongCat and Cai, Xunliang and Cheng, Meng and Gao, Feng and Kong, Zhe and Li, Jiamu and Li, Le and Li, Weiheng and Liu, Hongyu and Tan, Shuai and others},
  journal={arXiv preprint arXiv:2605.26486},
  year={2026}
}

@inproceedings{duan2025worldscore,
  title={Worldscore: A unified evaluation benchmark for world generation},
  author={Duan, Haoyi and Yu, Hong-Xing and Chen, Sirui and Fei-Fei, Li and Wu, Jiajun},
  booktitle={Proceedings of the IEEE/CVF international conference on computer vision},
  pages={27713--27724},
  year={2025}
}

@article{bai2026geovideo,
  title={Geovideo: Introducing geometric regularization into video generation model},
  author={Bai, Yunpeng and Fang, Shaoheng and Yu, Chaohui and Wang, Fan and Huang, Qixing},
  journal={Advances in Neural Information Processing Systems},
  volume={38},
  pages={57602--57622},
  year={2026}
}

@article{liu2026improving,
  title={Improving video generation with human feedback},
  author={Liu, Jie and Liu, Gongye and Liang, Jiajun and Yuan, Ziyang and Liu, Xiaokun and Zheng, Mingwu and Wu, Xiele and Wang, Qiulin and Xia, Menghan and Wang, Xintao and others},
  journal={Advances in Neural Information Processing Systems},
  volume={38},
  pages={82155--82192},
  year={2026}
}

@inproceedings{ma2025hpsv3,
  title={Hpsv3: Towards wide-spectrum human preference score},
  author={Ma, Yuhang and Wu, Xiaoshi and Sun, Keqiang and Li, Hongsheng},
  booktitle={2025 IEEE/CVF International Conference on Computer Vision (ICCV)},
  pages={15086--15095},
  year={2025},
  organization={IEEE}
}

@article{gu2026worldcycle,
  title={WorldCycle: Self-Verifiable Reinforcement Learning for Long-Horizon Video World Models},
  author={Gu, Bohai and Yuan, Yueyang and Wu, Taiyi and Du, Dazhao and Liu, Jian and Pang, Xiaoyi and Zhang, Jie and Lu, Xiaocheng and Zhong, Haobin and Zhao, Xiaotong and others},
  journal={arXiv preprint arXiv:2608.04964},
  year={2026}
}

@inproceedings{wang2025vggt,
  title={Vggt: Visual geometry grounded transformer},
  author={Wang, Jianyuan and Chen, Minghao and Karaev, Nikita and Vedaldi, Andrea and Rupprecht, Christian and Novotny, David},
  booktitle={2025 IEEE/CVF Conference on Computer Vision and Pattern Recognition (CVPR)},
  pages={5294--5306},
  year={2025},
  organization={IEEE}
}

@article{lin2025depth,
  title={Depth anything 3: Recovering the visual space from any views},
  author={Lin, Haotong and Chen, Sili and Liew, Junhao and Chen, Donny Y and Li, Zhenyu and Shi, Guang and Feng, Jiashi and Kang, Bingyi},
  journal={arXiv preprint arXiv:2511.10647},
  year={2025}
}

@article{lingbot-world-v2,
      title={Infinite Worlds with Versatile Interactions}, 
      author={Zelin Gao and Qiuyu Wang and Jiapeng Zhu and Jingye Chen and Zichen Liu and Qingyan Bai and Jiahao Wang and Yufeng Yuan and Hanlin Wang and Yichong Lu and Ka Leong Cheng and Haojie Zhang and Jian Gao and Tianrui Feng and Yuzheng Liu and Yao Yao and Yinghao Xu and Xing Zhu and Yujun Shen and Hao Ouyang},
      journal={arXiv preprint arXiv:2607.07534},
      year={2026}
}

@inproceedings{wang2026worldr1,
  title={World-R1: Reinforcing 3D Constraints for Text-to-Video Generation},
  author={Wang, Weijie and He, Xiaoxuan and Gu, Youping and Zhang, Zeyu and He, Yefei and Ding, Yanbo and Hu, Xirui and Chen, Donny Y. and He, Zhiyuan and Yang, Yuqing and Yang, Yifan and Zhuang, Bohan},
  booktitle={ICML},
  year={2026}
}

@misc{zhou2026uniworldview,
      title={UniWorld-View: Large-Baseline View Synthesis via Video Diffusion Models}, 
      author={Haiyang Zhou and Wangbo Yu and Chaoran Feng and Xunyu Zhou and Yonghong Tian and Li Yuan},
      year={2026},
      eprint={2608.04701},
      archivePrefix={arXiv},
      primaryClass={cs.CV},
      url={https://arxiv.org/abs/2608.04701}, 
}

@article{shen2026lyra2,
  title={Lyra 2.0: Explorable generative 3d worlds},
  author={Shen, Tianchang and Bahmani, Sherwin and He, Kai and Srinivasan, Sangeetha Grama and Cao, Tianshi and Ren, Jiawei and Li, Ruilong and Wang, Zian and Sharp, Nicholas and Gojcic, Zan and others},
  journal={arXiv preprint arXiv:2604.13036},
  year={2026}
}

@article{liu2026flow,
  title={Flow-grpo: Training flow matching models via online rl},
  author={Liu, Jie and Liu, Gongye and Liang, Jiajun and Li, Yangguang and Liu, Jiaheng and Wang, Xintao and Wan, Pengfei and Zhang, Di and Ouyang, Wanli},
  journal={Advances in neural information processing systems},
  volume={38},
  pages={40783--40818},
  year={2026}
}

@inproceedings{clark2024directly,
  title={Directly fine-tuning diffusion models on differentiable rewards},
  author={Clark, Kevin and Vicol, Paul and Swersky, Kevin and Fleet, David},
  booktitle={International Conference on Learning Representations},
  volume={2024},
  pages={4793--4822},
  year={2024}
}

@inproceedings{zheng2026diffusionnft,
  title={Diffusionnft: Online diffusion reinforcement with forward process},
  author={Zheng, Kaiwen and Chen, Huayu and Ye, Haotian and Wang, Haoxiang and Zhang, Qinsheng and Jiang, Kai and Su, Hang and Ermon, Stefano and Zhu, Jun and Liu, Ming-Yu},
  booktitle={International Conference on Learning Representations},
  volume={2026},
  pages={134129--134150},
  year={2026}
}

@article{rakheja2025world,
  title={World consistency score: a unified metric for video generation quality},
  author={Rakheja, Akshat and Ashdhir, Aarsh and Bhattacharjee, Aryan and Sharma, Vanshika},
  journal={arXiv preprint arXiv:2508.00144},
  year={2025}
}

@article{ye2026mind,
  title={Mind: Benchmarking memory consistency and action control in world models},
  author={Ye, Yixuan and Lu, Xuanyu and Jiang, Yuxin and Gu, Yuchao and Zhao, Rui and Liang, Qiwei and Pan, Jiachun and Zhang, Fengda and Wu, Weijia and Wang, Alex Jinpeng},
  journal={arXiv preprint arXiv:2602.08025},
  year={2026}
}

@article{rafailov2023direct,
  title={Direct preference optimization: Your language model is secretly a reward model},
  author={Rafailov, Rafael and Sharma, Archit and Mitchell, Eric and Manning, Christopher D and Ermon, Stefano and Finn, Chelsea},
  journal={Advances in neural information processing systems},
  volume={36},
  pages={53728--53741},
  year={2023}
}

@article{kerbl20233d,
  title={3d gaussian splatting for real-time radiance field rendering.},
  author={Kerbl, Bernhard and Kopanas, Georgios and Leimk{\"u}hler, Thomas and Drettakis, George and others},
  journal={ACM Trans. Graph.},
  volume={42},
  number={4},
  pages={139--1},
  year={2023}
}

@article{sun2025worldplay,
  title={Worldplay: Towards long-term geometric consistency for real-time interactive world modeling},
  author={Sun, Wenqiang and Zhang, Haiyu and Wang, Haoyuan and Wu, Junta and Wang, Zehan and Wang, Zhenwei and Wang, Yunhong and Zhang, Jun and Wang, Tengfei and Guo, Chunchao},
  journal={arXiv preprint arXiv:2512.14614},
  year={2025}
}

@inproceedings{wang2026taming,
  title={Taming camera-controlled video generation with verifiable geometry reward},
  author={Wang, Zhaoqing and Xia, Xiaobo and Bie, Zhuolin and Liu, Jinlin and Yu, Dongdong and Bian, Jia-Wang and Wang, Changhu},
  booktitle={European Conference on Computer Vision},
  pages={625--642},
  year={2026},
  organization={Springer}
}

@article{ge2026campilot,
  title={Campilot: Improving camera control in video diffusion model with efficient camera reward feedback},
  author={Ge, Wenhang and Shen, Guibao and Feng, Jiawei and Wang, Luozhou and Lu, Hao and Tian, Xingye and Tao, Xin and Chen, Ying-Cong},
  journal={arXiv preprint arXiv:2601.16214},
  year={2026}
}

@article{team2026dreamx,
  title={DreamX-World 1.0: A General-Purpose Interactive World Model},
  author={Team, DreamX and Bai, Yancheng and Chen, Rui and Chu, Xiangxiang and Dang, Rujing and Dou, Hao and Gao, Bingjie and Gu, Qiwen and Hong, Siyu and Lei, Jiachen and others},
  journal={arXiv preprint arXiv:2606.16993},
  year={2026}
}

@article{xu2024camco,
  title={Camco: Camera-controllable 3d-consistent image-to-video generation},
  author={Xu, Dejia and Nie, Weili and Liu, Chao and Liu, Sifei and Kautz, Jan and Wang, Zhangyang and Vahdat, Arash},
  journal={arXiv preprint arXiv:2406.02509},
  year={2024}
}

@inproceedings{ren2025gen3c,
  title={Gen3c: 3d-informed world-consistent video generation with precise camera control},
  author={Ren, Xuanchi and Shen, Tianchang and Huang, Jiahui and Ling, Huan and Lu, Yifan and Nimier-David, Merlin and M{\"u}ller, Thomas and Keller, Alexander and Fidler, Sanja and Gao, Jun},
  booktitle={2025 IEEE/CVF Conference on Computer Vision and Pattern Recognition (CVPR)},
  pages={6121--6132},
  year={2025},
  organization={IEEE}
}

@article{xiang2026videoweave,
  title={VideoWeave: Unlocking Geometric Consistency in Video Generation via Joint Geometry-Video Modeling},
  author={Xiang, Xunzhi and Duan, Zixuan and Chen, Yabo and Wei, Zhengxuan and Zhang, Guiyu and Gu, Zixiao and Gao, Zhe and Huang, Haibin and Zhang, Chi and Fan, Qi and others},
  journal={arXiv preprint arXiv:2606.14162},
  year={2026}
}

@inproceedings{zhuang2026workforceagent,
  title={WorkForceAgent-R1: Incentivizing reasoning capability in llm-based web agents via reinforcement learning},
  author={Zhuang, Yuchen and Jin, Di and Chen, Jiaao and Shi, Wenqi and Wang, Hanrui and Zhang, Chao},
  booktitle={Findings of the Association for Computational Linguistics: EACL 2026},
  pages={34--49},
  year={2026}
}

@misc{pexels,
  author       = {{Pexels}},
  title        = {Pexels: Free Stock Photos},
  title = {Pexels},
  year = {2026},
  url = {https://www.pexels.com},
  note = {Accessed: 2026-09-17}
}

@misc{google_nano_banana,
  author       = {Google},
  title        = {Nano Banana},
  year         = {2026},
  url          = {https://aistudio.google.com/docs/image-generation}
}

@inproceedings{ling2024dl3dv,
  title={Dl3dv-10k: A large-scale scene dataset for deep learning-based 3d vision},
  author={Ling, Lu and Sheng, Yichen and Tu, Zhi and Zhao, Wentian and Xin, Cheng and Wan, Kun and Yu, Lantao and Guo, Qianyu and Yu, Zixun and Lu, Yawen and others},
  booktitle={2024 IEEE/CVF Conference on Computer Vision and Pattern Recognition (CVPR)},
  pages={22160--22169},
  year={2024},
  organization={IEEE}
}

@misc{yang2024qwen2technicalreport,
      title={Qwen2 Technical Report}, 
      author={An Yang and Baosong Yang and Binyuan Hui and Bo Zheng and Bowen Yu and Chang Zhou and Chengpeng Li and Chengyuan Li and Dayiheng Liu and Fei Huang and Guanting Dong and Haoran Wei and Huan Lin and Jialong Tang and Jialin Wang and Jian Yang and Jianhong Tu and Jianwei Zhang and Jianxin Ma and Jianxin Yang and Jin Xu and Jingren Zhou and Jinze Bai and Jinzheng He and Junyang Lin and Kai Dang and Keming Lu and Keqin Chen and Kexin Yang and Mei Li and Mingfeng Xue and Na Ni and Pei Zhang and Peng Wang and Ru Peng and Rui Men and Ruize Gao and Runji Lin and Shijie Wang and Shuai Bai and Sinan Tan and Tianhang Zhu and Tianhao Li and Tianyu Liu and Wenbin Ge and Xiaodong Deng and Xiaohuan Zhou and Xingzhang Ren and Xinyu Zhang and Xipin Wei and Xuancheng Ren and Xuejing Liu and Yang Fan and Yang Yao and Yichang Zhang and Yu Wan and Yunfei Chu and Yuqiong Liu and Zeyu Cui and Zhenru Zhang and Zhifang Guo and Zhihao Fan},
      year={2024},
      eprint={2407.10671},
      archivePrefix={arXiv},
      primaryClass={cs.CL},
      url={https://arxiv.org/abs/2407.10671}, 
}

@inproceedings{li2023blip,
  title={Blip-2: Bootstrapping language-image pre-training with frozen image encoders and large language models},
  author={Li, Junnan and Li, Dongxu and Savarese, Silvio and Hoi, Steven},
  booktitle={International Conference on Machine Learning},
  pages={19730--19742},
  year={2023},
  organization={PMLR}
}

@InProceedings{dart_2026_CVPR,
    author    = {Zhang, Chuhan and Le Moing, Guillaume and Koppula, Skanda and Rocco, Ignacio and Momeni, Liliane and Xie, Junyu and Sun, Shuyang and Sukthankar, Rahul and Barral, Jo\"elle K. and Hadsell, Raia and Ghahramani, Zoubin and Zisserman, Andrew and Zhang, Junlin and Sajjadi, Mehdi S. M.},
    title     = {Efficiently Reconstructing Dynamic Scenes One D4RT at a Time},
    booktitle = {Proceedings of the IEEE/CVF Conference on Computer Vision and Pattern Recognition (CVPR)},
    month     = {June},
    year      = {2026},
    pages     = {7382-7392}
}

@inproceedings{schoenberger2016sfm,
    author={Sch\"{o}nberger, Johannes Lutz and Frahm, Jan-Michael},
    title={Structure-from-Motion Revisited},
    booktitle={Conference on Computer Vision and Pattern Recognition (CVPR)},
    year={2016},
}
\bibliographystyle{iclr2027_conference}

\clearpage
\appendix
\section{Appendix}
In the Appendix, we include additional explanation of the localized reward calculation, additional experimental result tables, additional training, evaluation, and benchmark construction details, as well as additional ablations and analyses.

\subsection{RL reward formulation with \method}

We instantiate voxelized credit assignment by making the advantage calculation non-uniform across a sample. For example, for DiffusionNFT, the canonical implementation is using a scalar $r$ with 
\begin{align}
\nonumber \mathcal{L}_{\mathrm{NFT}}  = \mathbb{E}\Big[r\|v_\theta^{+}-v\|^2 + (1-r)\|v_\theta^{-}-v\|^2\Big]
\end{align}
With \method, we assign a different $r$ per latent spatio-temporal patch by averaging the optimality probability (normalized reward) for all voxels that project to this patch:
\begin{align}
\mathcal{L}_{\mathrm{NFT},j}^{(k)} = \frac{1}{\vert A\vert}\mathbb{E}\Big[\sum_{i\in A = \{i: S_i \cap I_j(k) \neq \emptyset\}} r_i\|v_{\theta^{+},j}^{(k)}-v_j^{(k)}\|^2 + 
(1-r_i)\|v_{\theta^{-},j}^{(k)}-v_j^{(k)}\|^2\Big]
\end{align}
where $S_i$ denotes all pixels that project to voxel $i$, and $I_j(k)$ is the set of pixels in frame j patch k, and A denotes all voxels that project onto the current patch. The FlowGRPO~\citep{liu2026flow} objective can be decomposed similarly. 

\subsection{Additional experiment results}
We provide full results referenced in the main text below. We present the full  easy/medium/hard breakdown for TrajectoryBench evaluation for all metrics. We also show standard errors to demonstrate the statistical significance of metric improvements.

Table~\ref{tab:trajbenchfull} compares \method with baselines, with full breakdown of easy/medium/hard split of all 3D metrics and standard errors on TrajectoryBench.
Table~\ref{tab:dl3dv_se} shows \method's comparison with baselines on DL3DV with standard errors.
Table~\ref{tab:ablation_ca_full} shows full difficulty breakdown for the ablation of different localization methods, demonstrating that voxel localization (used in \method) is the strongest, with growing advantage as the difficulty increases.
Table~\ref{tab:ablation_rl_full} shows full difficulty breakdown of the \method evaluation with various post-training algorithms, including DiffusionNFT, Flow-GRPO, and DRaFT, with standard errors.
Table~\ref{tab:ablation_voxel_size} shows various metrics' improvement under different voxel granularities for voxelization.
Table~\ref{tab:ablation_recipe_se} shows standard errors for the \method recipe ablation.

\begin{table*}[h!]
\centering
\small
\setlength{\tabcolsep}{4pt}
\newcommand{\std}[1]{{\tiny$\pm$#1}}
\resizebox{\linewidth}{!}{
\begin{tabular}{@{}ll *{18}{c}@{}}
\toprule
& \multirow{3}{*}{Model}
 & \multicolumn{15}{c}{3D consistency}
  & \multicolumn{2}{c}{Camera}
  & \multicolumn{1}{c}{Video quality} \\
 \cmidrule(lr){3-17} \cmidrule(lr){18-19} \cmidrule(lr){20-20}
 &  & \multicolumn{3}{c}{PSNR-V $\uparrow$} & \multicolumn{3}{c}{PSNR-D $\uparrow$} & \multicolumn{3}{c}{PSNR-GS $\uparrow$} & \multicolumn{3}{c}{MVCS $\uparrow$} & \multicolumn{3}{c}{Epipolar $\downarrow$} & \multirow{2}{*}{RPE$^\circ$ $\downarrow$} & \multirow{2}{*}{RPE trans $\downarrow$} & \multirow{2}{*}{VR-VQ $\uparrow$} \\
 \cmidrule(lr){3-5} \cmidrule(lr){6-8} \cmidrule(lr){9-11} \cmidrule(lr){12-14} \cmidrule(lr){15-17}
 & & E & M & H & E & M & H & E & M & H & E & M & H & E & M & H & & & \\
\midrule
\multirow{5}{*}{Lingbot2} & Base & 22.6 & 18.2 & 15.3 & 20.6 & 17.6 & 14.5 & 21.0 & 18.0 & 15.7 & 0.918 & 0.878 & 0.812 & 0.524 & 0.681 & 2.034 & 13.2 & 0.108 & 0.21\\
 & VideoGPA & 22.7\std{0.03} & 18.3\std{0.08} & 15.4\std{0.04} & 20.7\std{0.05} & 17.7\std{0.08} & 14.6\std{0.05} & 21.1\std{0.05} & 18.1\std{0.07} & 15.8\std{0.04} & 0.916\std{0.002} & 0.880\std{0.004} & 0.820\std{0.005} & 0.533\std{0.011} & 0.684\std{0.013} & 1.997\std{0.052} & 13.1\std{0.14} & 0.107\std{0.001} & 0.22\std{0.004}\\
 & World-R1 & 22.7\std{0.06} & 18.1\std{0.08} & 15.4\std{0.04} & 20.6\std{0.08} & 17.4\std{0.08} & 14.5\std{0.05} & 21.1\std{0.06} & 18.0\std{0.07} & 15.7\std{0.04} & 0.917\std{0.002} & 0.873\std{0.005} & 0.809\std{0.005} & 0.526\std{0.012} & 0.681\std{0.012} & 2.046\std{0.054} & 13.1\std{0.14} & 0.108\std{0.001} & 0.21\std{0.004}\\
 & Global (ours) & 22.9\std{0.14} & 19.3\std{0.11} & 16.5\std{0.05} & 20.8\std{0.13} & 18.8\std{0.11} & 15.7\std{0.06} & 21.4\std{0.11} & 18.9\std{0.09} & 16.7\std{0.05} & 0.916\std{0.005} & 0.908\std{0.006} & 0.848\std{0.005} & 0.521\std{0.013} & 0.649\std{0.015} & 1.726\std{0.057} & 11.2\std{0.17} & 0.102\std{0.001} & 0.31\std{0.006}\\
 & \method (ours) & \textbf{23.8\std{0.16}} & \textbf{20.2\std{0.11}} & \textbf{17.6\std{0.05}} & \textbf{21.6\std{0.15}} & \textbf{19.7\std{0.11}} & \textbf{17.0\std{0.06}} & \textbf{21.7\std{0.11}} & \textbf{19.4\std{0.09}} & \textbf{17.4\std{0.05}} & \textbf{0.942\std{0.006}} & \textbf{0.929\std{0.006}} & \textbf{0.871\std{0.006}} & \textbf{0.490\std{0.010}} & \textbf{0.601\std{0.012}} & \textbf{1.439\std{0.052}} & 8.9\std{0.18} & 0.103\std{0.001} & 0.37\std{0.006}\\
\midrule
\multirow{5}{*}{UniWorld} & Base & 23.0 & 17.2 & 16.7 & 21.2 & 16.4 & 16.1 & 20.6 & 17.5 & 16.9 & 0.900 & 0.898 & 0.871 & 0.450 & \textbf{0.760} & 1.623 & 13.8 & 0.132 & 0.39\\
 & VideoGPA & 23.1\std{0.02} & 17.2\std{0.02} & 16.8\std{0.02} & 21.2\std{0.05} & 16.4\std{0.04} & 16.2\std{0.03} & 20.6\std{0.05} & 17.6\std{0.03} & 16.9\std{0.02} & 0.900\std{0.001} & 0.898\std{0.001} & 0.871\std{0.002} & 0.450\std{0.002} & 0.773\std{0.010} & 1.642\std{0.036} & 13.8\std{0.06} & 0.132\std{0.000} & 0.39\std{0.002}\\
 & World-R1 & 22.4\std{0.07} & 17.3\std{0.08} & 17.2\std{0.06} & 20.5\std{0.08} & 16.5\std{0.08} & 16.6\std{0.06} & 20.1\std{0.09} & 17.9\std{0.08} & 18.0\std{0.05} & 0.894\std{0.003} & 0.883\std{0.006} & 0.858\std{0.005} & 0.462\std{0.003} & 0.860\std{0.015} & 1.805\std{0.042} & 13.1\std{0.19} & 0.129\std{0.001} & 0.22\std{0.006}\\
 & Global (ours) & 22.8\std{0.05} & 17.6\std{0.05} & 17.6\std{0.05} & 20.8\std{0.06} & 16.9\std{0.07} & 17.1\std{0.06} & 20.5\std{0.07} & 17.9\std{0.06} & 17.6\std{0.04} & 0.899\std{0.001} & 0.901\std{0.005} & 0.887\std{0.003} & 0.447\std{0.003} & 0.773\std{0.017} & 1.514\std{0.039} & 13.3\std{0.14} & 0.126\std{0.001} & 0.37\std{0.004}\\
 & \method (ours) & \textbf{23.2\std{0.05}} & \textbf{18.2\std{0.06}} & \textbf{18.1\std{0.06}} & \textbf{21.3\std{0.07}} & \textbf{17.4\std{0.07}} & \textbf{17.5\std{0.06}} & \textbf{21.0\std{0.09}} & \textbf{18.5\std{0.06}} & \textbf{18.2\std{0.05}} & \textbf{0.904\std{0.002}} & \textbf{0.913\std{0.004}} & \textbf{0.895\std{0.003}} & \textbf{0.446\std{0.002}} & \textbf{0.760\std{0.012}} & \textbf{1.379\std{0.036}} & 14.0\std{0.18} & 0.130\std{0.001} & 0.41\std{0.004}\\
\midrule
\multirow{5}{*}{Lyra2} & Base & 22.2 & 17.9 & 17.8 & 21.1 & 17.2 & 17.3 & 21.6 & 18.9 & 18.5 & \textbf{0.924} & 0.922 & 0.881 & 0.519 & 0.874 & 2.027 & 7.9 & 0.100 & -0.23\\
 & VideoGPA & 22.1\std{0.04} & 17.7\std{0.06} & 17.3\std{0.04} & 21.1\std{0.05} & 17.1\std{0.06} & 16.7\std{0.04} & 21.6\std{0.06} & 18.7\std{0.06} & 18.0\std{0.04} & 0.920\std{0.003} & 0.926\std{0.003} & 0.880\std{0.003} & 0.519\std{0.003} & 0.878\std{0.033} & 1.963\std{0.050} & 7.7\std{0.11} & 0.099\std{0.001} & -0.21\std{0.003}\\
 & World-R1 & 22.2\std{0.05} & 18.1\std{0.06} & 18.4\std{0.04} & 21.0\std{0.07} & 17.5\std{0.06} & 17.8\std{0.04} & 21.7\std{0.07} & 19.3\std{0.08} & 19.0\std{0.04} & 0.920\std{0.003} & 0.929\std{0.003} & 0.883\std{0.003} & 0.529\std{0.007} & 0.939\std{0.034} & 1.999\std{0.051} & 7.7\std{0.12} & 0.101\std{0.001} & -0.20\std{0.003}\\
 & Global (ours) & \textbf{22.5\std{0.05}} & 18.6\std{0.06} & 19.0\std{0.04} & \textbf{21.3\std{0.07}} & 17.9\std{0.06} & 18.4\std{0.04} & \textbf{21.8\std{0.07}} & 19.6\std{0.07} & 19.4\std{0.04} & 0.921\std{0.003} & 0.929\std{0.003} & 0.884\std{0.003} & 0.513\std{0.003} & 0.869\std{0.023} & 1.784\std{0.050} & 7.6\std{0.11} & 0.099\std{0.001} & -0.21\std{0.003}\\
 & \method (ours) & \textbf{22.5\std{0.04}} & \textbf{18.8\std{0.06}} & \textbf{19.4\std{0.04}} & 21.2\std{0.06} & \textbf{18.1\std{0.07}} & \textbf{18.8\std{0.05}} & \textbf{21.8\std{0.08}} & \textbf{19.8\std{0.07}} & \textbf{19.7\std{0.04}} & 0.922\std{0.003} & \textbf{0.930\std{0.003}} & \textbf{0.892\std{0.003}} & \textbf{0.510\std{0.003}} & \textbf{0.847\std{0.025}} & \textbf{1.723\std{0.051}} & 7.5\std{0.10} & 0.101\std{0.001} & -0.19\std{0.003}\\
\bottomrule
\end{tabular}
}
\caption{TrajectoryBench evaluation difficulty breakdown. Geometry columns are split by difficulty
tier E / M / H for easy, medium, hard splits. \method achieves better 3D consistency than baseline methods in all twelve metrics across three splits for all three models, while maintaining or improving camera control and video quality, with the strongest advantage on the hard split. $\pm$ denotes paired standard error.}
\label{tab:trajbenchfull}
\end{table*}

\begin{table*}[h!]
    \centering
    \small
    \setlength{\tabcolsep}{4pt}
    \newcommand{\std}[1]{{\tiny$\pm$#1}}
    \resizebox{\linewidth}{!}{
    \begin{tabular}{@{}ll *{8}{c}@{}}
    \toprule
    & \multirow{2}{*}{Model}
     & \multicolumn{5}{c}{3D consistency}
      & \multicolumn{2}{c}{Camera}
      & \multicolumn{1}{c}{Video quality} \\
     \cmidrule(lr){3-7} \cmidrule(lr){8-9} \cmidrule(lr){10-10}
     & & PSNR-V $\uparrow$ & PSNR-D $\uparrow$ & PSNR-GS $\uparrow$ & MVCS $\uparrow$ & Epipolar $\downarrow$ & RPE$^\circ$ $\downarrow$ & RPE trans $\downarrow$ & VR-VQ $\uparrow$ \\
    \midrule
    \multirow{5}{*}{Lingbot2} & Base & 13.3 & 13.1 & 15.6 & 0.719 & 2.161 & 20.52 & 0.082 & -0.422\\
 & VideoGPA & 13.3\std{0.10} & 13.2\std{0.11} & 15.6\std{0.09} & 0.728\std{0.014} & 2.429\std{0.158} & 20.15\std{1.215} & 0.081\std{0.005} & -0.409\std{0.010}\\
 & World-R1 & 13.6\std{0.11} & 13.3\std{0.10} & 15.7\std{0.11} & 0.717\std{0.017} & 2.271\std{0.154} & 20.13\std{1.681} & 0.070\std{0.004} & -0.429\std{0.012}\\
 & Global (ours) & 14.0\std{0.15} & 13.8\std{0.14} & 16.3\std{0.14} & 0.772\std{0.016} & 2.107\std{0.152} & 22.68\std{1.745} & 0.086\std{0.006} & -0.305\std{0.017}\\
 & \method{} (ours) & \textbf{16.1\std{0.22}} & \textbf{15.8\std{0.20}} & \textbf{17.3\std{0.21}} & \textbf{0.795\std{0.021}} & 1.365\std{0.198} & 18.33\std{1.504} & 0.077\std{0.006} & -0.283\std{0.018}\\
\cmidrule(lr){1-10}
\multirow{5}{*}{UniWorld} & Base & 14.7 & 14.5 & 16.6 & 0.748 & 1.510 & 12.94 & 0.057 & -0.128\\
 & VideoGPA & 14.7\std{0.03} & 14.6\std{0.07} & 16.7\std{0.05} & 0.752\std{0.005} & 1.623\std{0.082} & 13.31\std{0.458} & 0.058\std{0.001} & -0.129\std{0.008}\\
 & World-R1 & 14.9\std{0.12} & 14.7\std{0.12} & 17.3\std{0.14} & 0.747\std{0.018} & 1.926\std{0.102} & 15.45\std{1.432} & 0.064\std{0.002} & -0.234\std{0.015}\\
 & Global (ours) & 14.9\std{0.10} & 14.7\std{0.11} & 16.7\std{0.07} & 0.778\std{0.013} & 1.557\std{0.079} & 12.18\std{0.514} & 0.058\std{0.002} & -0.134\std{0.011}\\
 & \method{} (ours) & \textbf{15.4\std{0.11}} & \textbf{15.1\std{0.12}} & \textbf{17.4\std{0.09}} & \textbf{0.807\std{0.014}} & 1.429\std{0.072} & 12.64\std{0.949} & 0.055\std{0.002} & -0.121\std{0.012}\\
\cmidrule(lr){1-10}
\multirow{5}{*}{Lyra2} & Base & 15.2 & 14.8 & 16.8 & 0.819 & 2.448 & 22.72 & 0.148 & -0.557\\
 & VideoGPA & 15.0\std{0.09} & 14.7\std{0.08} & 16.6\std{0.09} & 0.813\std{0.011} & 2.595\std{0.216} & 21.95\std{0.549} & 0.143\std{0.003} & -0.508\std{0.011}\\
 & World-R1 & 15.3\std{0.11} & 15.2\std{0.13} & 16.9\std{0.12} & 0.832\std{0.010} & 2.213\std{0.200} & 21.72\std{0.768} & 0.140\std{0.003} & -0.523\std{0.011}\\
 & Global (ours) & 15.6\std{0.11} & 15.3\std{0.13} & 17.1\std{0.10} & 0.826\std{0.013} & 2.244\std{0.253} & 22.17\std{0.619} & 0.145\std{0.003} & -0.525\std{0.010}\\
 & \method{} (ours) & \textbf{15.9\std{0.12}} & \textbf{15.6\std{0.13}} & \textbf{17.3\std{0.10}} & \textbf{0.844\std{0.012}} & 1.863\std{0.225} & 22.38\std{0.690} & 0.145\std{0.002} & -0.513\std{0.012}\\
\bottomrule
    \end{tabular}
    }
    \caption{DL3DV evaluation with standard errors. $\pm$ denotes paired standard errors. \method yields more improvement in 3D consistency compared to baseline post-training methods on three different base models for all four metrics, while maintaining camera control and video
    quality. The improvement is significant, with up to $37\%$ decrease in epipolar error.}
    \label{tab:dl3dv_se}
    \end{table*}
    
\begin{table}[h!]
\centering
\small
\setlength{\tabcolsep}{4pt}
\newcommand{\std}[1]{{\tiny$\pm$#1}}
\resizebox{\linewidth}{!}{
\begin{tabular}{@{}l *{18}{c}@{}}
\toprule
\multirow{2}{*}{Reward readout} & \multicolumn{15}{c}{3D consistency}
  & \multicolumn{2}{c}{Camera} & \multicolumn{1}{c}{Video quality}\\
 \cmidrule(lr){2-16} \cmidrule(lr){17-18} \cmidrule(lr){19-19}
& \multicolumn{3}{c}{PSNR-V $\uparrow$} & \multicolumn{3}{c}{PSNR-D $\uparrow$} & \multicolumn{3}{c}{PSNR-GS $\uparrow$} & \multicolumn{3}{c}{MVCS $\uparrow$} & \multicolumn{3}{c}{Epipolar $\downarrow$}
  & \multirow{2}{*}{RPE$^\circ$ $\downarrow$} & \multirow{2}{*}{RPE trans $\downarrow$}
  & \multirow{2}{*}{VR-VQ $\uparrow$}\\
 \cmidrule(lr){2-4} \cmidrule(lr){5-7} \cmidrule(lr){8-10} \cmidrule(lr){11-13} \cmidrule(lr){14-16}
& E & M & H & E & M & H & E & M & H & E & M & H & E & M & H & & &\\
\midrule
Base & 22.6 & 18.2 & 15.3 & 20.6 & 17.6 & 14.5 & 21.0 & 18.0 & 15.7 & 0.918 & 0.878 & 0.812 & 0.524 & 0.681 & 2.034 & 13.2 & 0.108 & 0.21\\
Global & 22.9\std{0.14} & 19.3\std{0.11} & 16.5\std{0.05} & 20.8\std{0.13} & 18.8\std{0.11} & 15.7\std{0.06} & 21.4\std{0.11} & 18.9\std{0.09} & 16.7\std{0.05} & 0.916\std{0.005} & 0.908\std{0.006} & 0.848\std{0.005} & 0.521\std{0.013} & 0.649\std{0.015} & 1.726\std{0.057} & 11.2\std{0.17} & 0.102\std{0.001} & 0.31\std{0.006}\\
Window & 23.3\std{0.17} & 19.9\std{0.11} & 17.0\std{0.05} & 21.2\std{0.15} & 19.3\std{0.11} & 16.3\std{0.06} & 21.6\std{0.12} & \textbf{19.4\std{0.09}} & 17.1\std{0.05} & 0.918\std{0.005} & 0.914\std{0.006} & 0.856\std{0.005} & 0.539\std{0.020} & 0.635\std{0.014} & 1.666\std{0.053} & 9.5\std{0.17} & \textbf{0.094\std{0.001}} & 0.32\std{0.006}\\
Framewise & 23.5\std{0.17} & 19.7\std{0.11} & 17.0\std{0.05} & 21.3\std{0.16} & 19.0\std{0.11} & 16.4\std{0.06} & 21.5\std{0.11} & 19.0\std{0.09} & 16.9\std{0.05} & 0.929\std{0.005} & 0.918\std{0.006} & 0.857\std{0.005} & 0.531\std{0.021} & 0.617\std{0.013} & 1.531\std{0.053} & 10.7\std{0.17} & 0.101\std{0.001} & 0.24\std{0.006}\\
Patchwise & 23.4\std{0.16} & 19.6\std{0.11} & 16.9\std{0.06} & 21.2\std{0.15} & 18.8\std{0.11} & 16.2\std{0.06} & 21.5\std{0.12} & 19.0\std{0.09} & 17.0\std{0.05} & 0.930\std{0.005} & 0.918\std{0.006} & 0.865\std{0.005} & 0.509\std{0.010} & 0.621\std{0.013} & 1.660\std{0.054} & \textbf{8.4\std{0.18}} & 0.100\std{0.001} & 0.24\std{0.006}\\
Voxel & \textbf{23.9\std{0.14}} & \textbf{20.3\std{0.11}} & \textbf{17.8\std{0.05}} & \textbf{21.6\std{0.14}} & \textbf{19.7\std{0.11}} & \textbf{17.2\std{0.06}} & \textbf{21.7\std{0.11}} & \textbf{19.4\std{0.09}} & \textbf{17.7\std{0.05}} & \textbf{0.937\std{0.005}} & \textbf{0.934\std{0.006}} & \textbf{0.874\std{0.005}} & \textbf{0.499\std{0.029}} & \textbf{0.587\std{0.013}} & \textbf{1.381\std{0.053}} & 8.9\std{0.18} & 0.107\std{0.001} & \textbf{0.38\std{0.006}}\\
\bottomrule
\end{tabular}
}
\caption{Full ablation table of different ways to localize the reward. We compare voxel reward (used in \method as local reward), and other localization schemes such as framewise, 2D patchwise, frame windows (40f windows). Voxel localization yields the best improvement because it normalizes per-group rollouts in the native 3D space where 3D consistency is measured, without being affected by shifts in region represented per frame/patch caused by slightly different camera following in each rollout. This advantage persists across difficulty splits, and is most significant on the ``hard'' split, which has longer generation horizon and more complex camera motions.}
\label{tab:ablation_ca_full}
\end{table}

\begin{table}[h!]
\centering
\small
\setlength{\tabcolsep}{4pt}
\newcommand{\std}[1]{{\tiny$\pm$#1}}
\resizebox{\linewidth}{!}{
\begin{tabular}{@{}ll *{18}{c}@{}}
\toprule
 & \multirow{2}{*}{RL algorithm} & \multicolumn{15}{c}{3D consistency}
  & \multicolumn{2}{c}{Camera} & \multicolumn{1}{c}{Video quality}\\
 \cmidrule(lr){3-17} \cmidrule(lr){18-19} \cmidrule(lr){20-20}
 & & \multicolumn{3}{c}{PSNR-V $\uparrow$} & \multicolumn{3}{c}{PSNR-D $\uparrow$} & \multicolumn{3}{c}{PSNR-GS $\uparrow$} & \multicolumn{3}{c}{MVCS $\uparrow$} & \multicolumn{3}{c}{Epipolar $\downarrow$}
  & \multirow{2}{*}{RPE$^\circ$ $\downarrow$} & \multirow{2}{*}{RPE trans $\downarrow$}
  & \multirow{2}{*}{VR-VQ $\uparrow$}\\
 \cmidrule(lr){3-5} \cmidrule(lr){6-8} \cmidrule(lr){9-11} \cmidrule(lr){12-14} \cmidrule(lr){15-17}
 & & E & M & H & E & M & H & E & M & H & E & M & H & E & M & H & & &\\
\midrule
\multirow{1}{*}{} & Base & 23.0 & 17.2 & 16.7 & 21.2 & 16.4 & 16.1 & 20.6 & 17.5 & 16.9 & 0.900 & 0.898 & 0.871 & 0.450 & 0.760 & 1.623 & 13.8 & 0.132 & 0.39\\
\cmidrule(lr){1-20}
\multirow{2}{*}{DiffusionNFT} & Ours-global & 22.8\std{0.05} & 17.6\std{0.05} & 17.6\std{0.05} & 20.8\std{0.06} & 16.9\std{0.07} & 17.1\std{0.06} & 20.5\std{0.07} & 17.9\std{0.06} & 17.6\std{0.04} & 0.899\std{0.001} & 0.901\std{0.005} & 0.887\std{0.003} & 0.447\std{0.003} & \textbf{0.773\std{0.017}} & 1.514\std{0.039} & \textbf{13.3\std{0.14}} & \textbf{0.126\std{0.001}} & 0.37\std{0.004}\\
 & LoGlo & \textbf{23.1\std{0.05}} & \textbf{17.9\std{0.06}} & \textbf{17.9\std{0.06}} & \textbf{21.2\std{0.07}} & \textbf{17.2\std{0.07}} & \textbf{17.4\std{0.06}} & \textbf{20.9\std{0.08}} & \textbf{18.4\std{0.07}} & \textbf{18.0\std{0.05}} & \textbf{0.903\std{0.002}} & \textbf{0.912\std{0.004}} & \textbf{0.895\std{0.003}} & \textbf{0.446\std{0.003}} & 0.781\std{0.017} & \textbf{1.436\std{0.037}} & 13.9\std{0.18} & 0.130\std{0.001} & \textbf{0.40\std{0.004}}\\
\cmidrule(lr){1-20}
\multirow{2}{*}{FlowGRPO} & Ours-global & 23.2\std{0.03} & 17.7\std{0.03} & 17.5\std{0.03} & 21.3\std{0.06} & 16.8\std{0.04} & 16.8\std{0.04} & 20.9\std{0.10} & 18.0\std{0.04} & 17.5\std{0.03} & 0.901\std{0.001} & \textbf{0.902\std{0.002}} & 0.880\std{0.002} & 0.456\std{0.001} & 0.788\std{0.014} & \textbf{1.500\std{0.035}} & 13.8\std{0.07} & \textbf{0.132\std{0.000}} & \textbf{0.37\std{0.003}}\\
 & LoGlo & \textbf{23.3\std{0.03}} & \textbf{17.8\std{0.03}} & \textbf{17.6\std{0.03}} & \textbf{21.5\std{0.05}} & \textbf{17.0\std{0.04}} & \textbf{16.9\std{0.04}} & \textbf{21.0\std{0.08}} & \textbf{18.1\std{0.04}} & \textbf{17.6\std{0.03}} & \textbf{0.902\std{0.001}} & \textbf{0.902\std{0.002}} & \textbf{0.883\std{0.002}} & \textbf{0.454\std{0.002}} & \textbf{0.764\std{0.008}} & 1.553\std{0.039} & \textbf{13.7\std{0.10}} & \textbf{0.132\std{0.000}} & 0.36\std{0.003}\\
\cmidrule(lr){1-20}
\multirow{2}{*}{DRaFT} & Ours-global & \textbf{23.0\std{0.02}} & 17.1\std{0.02} & 16.6\std{0.02} & \textbf{21.1\std{0.05}} & 16.2\std{0.04} & 16.0\std{0.03} & \textbf{20.6\std{0.07}} & 17.4\std{0.03} & 16.7\std{0.02} & \textbf{0.899\std{0.000}} & \textbf{0.894\std{0.001}} & 0.869\std{0.002} & 0.453\std{0.004} & 0.768\std{0.011} & 1.726\std{0.037} & \textbf{14.0\std{0.07}} & \textbf{0.132\std{0.000}} & \textbf{0.39\std{0.003}}\\
 & LoGlo & \textbf{23.0\std{0.02}} & \textbf{17.2\std{0.02}} & \textbf{16.8\std{0.02}} & \textbf{21.1\std{0.04}} & \textbf{16.3\std{0.04}} & \textbf{16.2\std{0.03}} & \textbf{20.6\std{0.06}} & \textbf{17.5\std{0.03}} & \textbf{16.8\std{0.02}} & \textbf{0.899\std{0.001}} & 0.893\std{0.002} & \textbf{0.871\std{0.002}} & \textbf{0.449\std{0.002}} & \textbf{0.765\std{0.013}} & \textbf{1.670\std{0.038}} & \textbf{14.0\std{0.08}} & \textbf{0.132\std{0.000}} & \textbf{0.39\std{0.003}}\\
\bottomrule
\end{tabular}
}
\caption{Full difficulty breakdown of TrajectoryBench evaluation of \method with three different post-training algorithms: DiffusionNFT, Flow-GRPO, and DRaFT. \method's advantage persists regardless of post-training algorithms. Paired standard error is displayed in each cell.}
\label{tab:ablation_rl_full}
\end{table}

\begin{table}[h!]
\centering
\small
\setlength{\tabcolsep}{4pt}
\newcommand{\std}[1]{{\tiny$\pm$#1}}
\resizebox{\linewidth}{!}{
\begin{tabular}{@{}ll *{18}{c}@{}}
\toprule
 & \multirow{2}{*}{Voxel size} & \multicolumn{15}{c}{3D consistency}
  & \multicolumn{2}{c}{Camera} & Video quality\\
 \cmidrule(lr){3-17} \cmidrule(lr){18-19} \cmidrule(lr){20-20}
 & & \multicolumn{3}{c}{PSNR-V $\uparrow$} & \multicolumn{3}{c}{PSNR-D $\uparrow$} & \multicolumn{3}{c}{PSNR-GS $\uparrow$} & \multicolumn{3}{c}{MVCS $\uparrow$} & \multicolumn{3}{c}{Epipolar $\downarrow$}
   & \multirow{2}{*}{RPE$^\circ$ $\downarrow$} & \multirow{2}{*}{RPE trans $\downarrow$} & \multirow{2}{*}{VR-VQ $\uparrow$}\\
    \cmidrule(lr){3-5} \cmidrule(lr){6-8} \cmidrule(lr){9-11} \cmidrule(lr){12-14} \cmidrule(lr){15-17}
 & & E & M & H & E & M & H & E & M & H & E & M & H & E & M & H & & & \\
\midrule
\multirow{6}{*}{} & Base & 22.6 & 18.2 & 15.3 & 20.6 & 17.6 & 14.5 & 21.0 & 18.0 & 15.7 & 0.918 & 0.878 & 0.812 & 0.524 & 0.681 & 2.034 & 13.2 & 0.108 & 0.21\\
 & 0.05 & 24.1\std{0.16} & \textbf{20.7\std{0.11}} & 18.6\std{0.06} & 21.8\std{0.16} & 20.1\std{0.11} & 17.8\std{0.06} & 21.7\std{0.13} & \textbf{19.7\std{0.09}} & 18.2\std{0.06} & 0.932\std{0.005} & 0.935\std{0.007} & 0.880\std{0.005} & \textbf{0.466\std{0.010}} & 0.579\std{0.014} & \textbf{1.229\std{0.051}} & 9.2\std{0.17} & 0.108\std{0.001} & 0.42\std{0.006}\\
 & 0.1 & 24.3\std{0.15} & 20.5\std{0.11} & \textbf{18.7\std{0.06}} & 21.8\std{0.15} & 19.9\std{0.11} & \textbf{17.9\std{0.07}} & \textbf{21.8\std{0.10}} & 19.6\std{0.09} & \textbf{18.4\std{0.06}} & 0.941\std{0.005} & 0.939\std{0.007} & 0.885\std{0.006} & 0.486\std{0.014} & \textbf{0.577\std{0.014}} & 1.235\std{0.051} & \textbf{9.0\std{0.19}} & 0.111\std{0.001} & 0.42\std{0.006}\\
 & 0.3 & \textbf{24.6\std{0.16}} & \textbf{20.7\std{0.11}} & 18.4\std{0.06} & \textbf{22.2\std{0.16}} & \textbf{20.2\std{0.11}} & 17.6\std{0.07} & 21.7\std{0.11} & 19.6\std{0.09} & 18.0\std{0.06} & \textbf{0.947\std{0.006}} & \textbf{0.940\std{0.007}} & \textbf{0.893\std{0.006}} & 0.467\std{0.012} & 0.581\std{0.013} & 1.246\std{0.051} & 10.5\std{0.18} & 0.117\std{0.001} & 0.42\std{0.006}\\
 & 0.5 & 23.6\std{0.12} & 19.9\std{0.11} & 17.5\std{0.06} & 21.4\std{0.12} & 19.3\std{0.11} & 16.6\std{0.06} & 21.6\std{0.10} & 19.2\std{0.09} & 17.5\std{0.05} & 0.933\std{0.004} & 0.923\std{0.006} & 0.865\std{0.005} & 0.484\std{0.016} & 0.595\std{0.013} & 1.450\std{0.053} & 14.5\std{0.18} & 0.119\std{0.001} & 0.39\std{0.006}\\
 & 0.7 & 24.2\std{0.17} & 20.3\std{0.11} & 17.7\std{0.06} & 21.8\std{0.16} & 19.7\std{0.11} & 17.0\std{0.06} & 21.8\std{0.11} & 19.4\std{0.09} & 17.6\std{0.05} & 0.935\std{0.005} & 0.929\std{0.006} & 0.882\std{0.005} & 0.489\std{0.014} & 0.588\std{0.013} & 1.321\std{0.052} & 10.3\std{0.19} & \textbf{0.107\std{0.001}} & \textbf{0.43\std{0.006}}\\
\bottomrule
\end{tabular}
}
\caption{Ablation of localization voxel size on TrajectoryBench, with difficulty breakdown and paired standard errors. Decrease of voxel size generally increases 3D consistency, up to 0.1 $\times$ median scene depth. We choose voxel size of 0.1 based on this ablation.}
\label{tab:ablation_voxel_size}
\end{table}

\begin{table}[h!]
\centering
\small
\setlength{\tabcolsep}{4pt}
\newcommand{\std}[1]{{\tiny$\pm$#1}}
\resizebox{\linewidth}{!}{
\begin{tabular}{@{}l *{9}{c}@{}}
\toprule
\multirow{2}{*}{} & \multicolumn{5}{c}{3D consistency}
 & \multicolumn{2}{c}{Camera} & \multicolumn{2}{c}{Video quality}\\
 \cmidrule(lr){2-6} \cmidrule(lr){7-8} \cmidrule(lr){9-10}
 & PSNR-V $\uparrow$ & PSNR-D $\uparrow$ & PSNR-GS $\uparrow$ & MVCS $\uparrow$ & Epipolar $\downarrow$ & RPE$^\circ$ $\downarrow$ & RPE trans $\downarrow$ & VR-VQ $\uparrow$ & HPSv3 $\uparrow$\\
\midrule
Base & 16.9 & 16.0 & 16.9 & 0.842 & 1.507 & 13.2 & 0.108 & 0.21 & 5.09\\
Local reward & \textbf{19.2\std{0.05}} & 18.2\std{0.05} & \textbf{18.7\std{0.04}} & 0.893\std{0.004} & 1.078\std{0.034} & 11.2\std{0.12} & 0.116\std{0.001} & 0.32\std{0.004} & 4.47\std{0.019}\\
+ interleave & \textbf{19.2\std{0.05}} & \textbf{18.3\std{0.05}} & 18.6\std{0.04} & \textbf{0.897\std{0.004}} & \textbf{1.072\std{0.034}} & \textbf{8.9\std{0.13}} & 0.107\std{0.001} & \textbf{0.38\std{0.004}} & 4.94\std{0.013}\\
+ global (final) & 19.0\std{0.05} & 18.2\std{0.05} & 18.4\std{0.04} & 0.894\std{0.004} & 1.111\std{0.033} & \textbf{8.9\std{0.13}} & \textbf{0.103\std{0.001}} & 0.37\std{0.004} & \textbf{5.18\std{0.010}}\\
\bottomrule
\end{tabular}
}
\caption{Ablation on \method recipe with paired standard errors. Using local reward only yields best 3D consistency but worst camera and aesthetics metrics. Adding in reward interleaving improves other metrics, but still degrades HPSv3 compared to the base model. \method, which includes both reward interleaving and local-global blending, yields the best overall result.}
\label{tab:ablation_recipe_se}
\end{table}

\subsection{Additional training details}
\label{sec:additional_training_details}
We perform post-training with \method on Lyra2, Lingbot2, and UniWorld on 1700 DL3DV scenes. For training, we use DL3DV's given camera trajectories based on COLMAP~\citep{schoenberger2016sfm}. Since each model's translation unit is different from the COLMAP unit, we perform scaling of $0.4\times$ to convert to Lyra2 unit, and $1\times$ for UniWorld. Lingbot2 does not need scaling since it self-normalizes by each scene's peak step.
We subsample the first half of the scene trajectories into 240 frames to ensure appropriate scene coverage and camera motion speed. 
We train all models with a batch size of 8 and group size of 16. For Lingbot2 and UniWorld, we use a group size of 16 and learning rate of 3e-4. For Lyra2, we train with a learning rate 3.33e-5 for more stability.
For Lingbot2 and UniWorld, we use an equal weighting (0.5-0.5) of RGB and depth error in the reward. For Lyra2, we use a 0.3-RGB, 0.7-depth weighting to reduce reward hacking.
For UniWorld, since it uses BLIP-2~\citep{li2023blip} captioner for inference, we also train with BLIP-2 captions. We do not use captions for Lingbot2 and Lyra2.
Models are trained till the geometry reward converges, around 16-36 hours on 64 H100s for the three base models.

\subsection{Additional evaluation details}
For depth PSNR evaluation, we use leave-one-out reprojection: we fuse keyframes (every 20th frame) from the video, leaving out the current frame, and project the point cloud using the given frame's camera pose. Leave-one-out is used because for long trajectories, keyframes can be spatially sparse and overall point cloud reprojection could be dominated by each frame's self-reprojection if not excluded.

\subsection{Additional details on TrajectoryBench construction}
TrajectoryBench contains 2000 tasks, each composed of an initial image, a camera trajectory, and an optional text prompt. We divide the benchmark into easy, medium, and hard splits. The scene types include indoor, outdoor, and ``extension'': one space going through a door into another space. 

The easy split is a subsample of WorldScore of 80 frames and 1 camera action, sourced from a subset of WorldScore~\citep{duan2025worldscore} to be comparable with existing benchmarks, thus putting the more difficult tasks in context. While WorldScore has a small split of ``composite camera action'' evaluation, it tests I2V models via three separate 80-frame, single-camera-action generations. Our medium split includes indoor, outdoor, and ``extension'' scene types. For indoor and outdoor, tests 3-action composition with a single 240-frame generation using the same initial images for comparability. Additionally, the medium split also includes ``extension'' scenes which go through a door into a new space.
For the hard splits (up to 240 frames and 8 camera motions), we source initial images for the indoor and outdoor scenes from Pexels~\citep{pexels}. Indoor covers residential, commercial, institutional, industrial, recreational, cultural and transit spaces. Outdoor covers civic gathering, transportation, agricultural (farms, vineyards), commercial, recreational, and natural spaces. For the ``extension'' scene type, since we require the door to be of specific size and location, and want to maintain scene diversity, we leverage Nano Banana~\citep{google_nano_banana} to generate spaces covering diverse scene types, such as ballroom, indoor pool, aircraft hangar, and courtroom. The camera trajectories are human-designed using language primitives (go back, go left, etc.), mapped into camera paths, and evaluated on base models to adjust the translation and rotation amount to ensure the viewer 1) does not crash into walls or objects, and 2) cover a substantial sweep of spaces.

We also include stylized scenes: 20\% of the benchmark is stylized via Nano Banana into eight styles, including Japanese anime, classical oil painting, Minecraft, watercolor, geometric low poly 3D, 3D studio cartoon (Pixar style), cyberpunk, and Victorian steampunk. The diverse coverage of TrajectoryBench and the inclusion of various difficulty levels of trajectories make enable a comprehensive evaluation of models.

\subsection{Additional Analyses and Ablations}
\label{sec:additional}

\subsubsection{Additional visualization of \method's local reward.}
\begin{figure}
  \centering
  \includegraphics[width=0.6\textwidth]{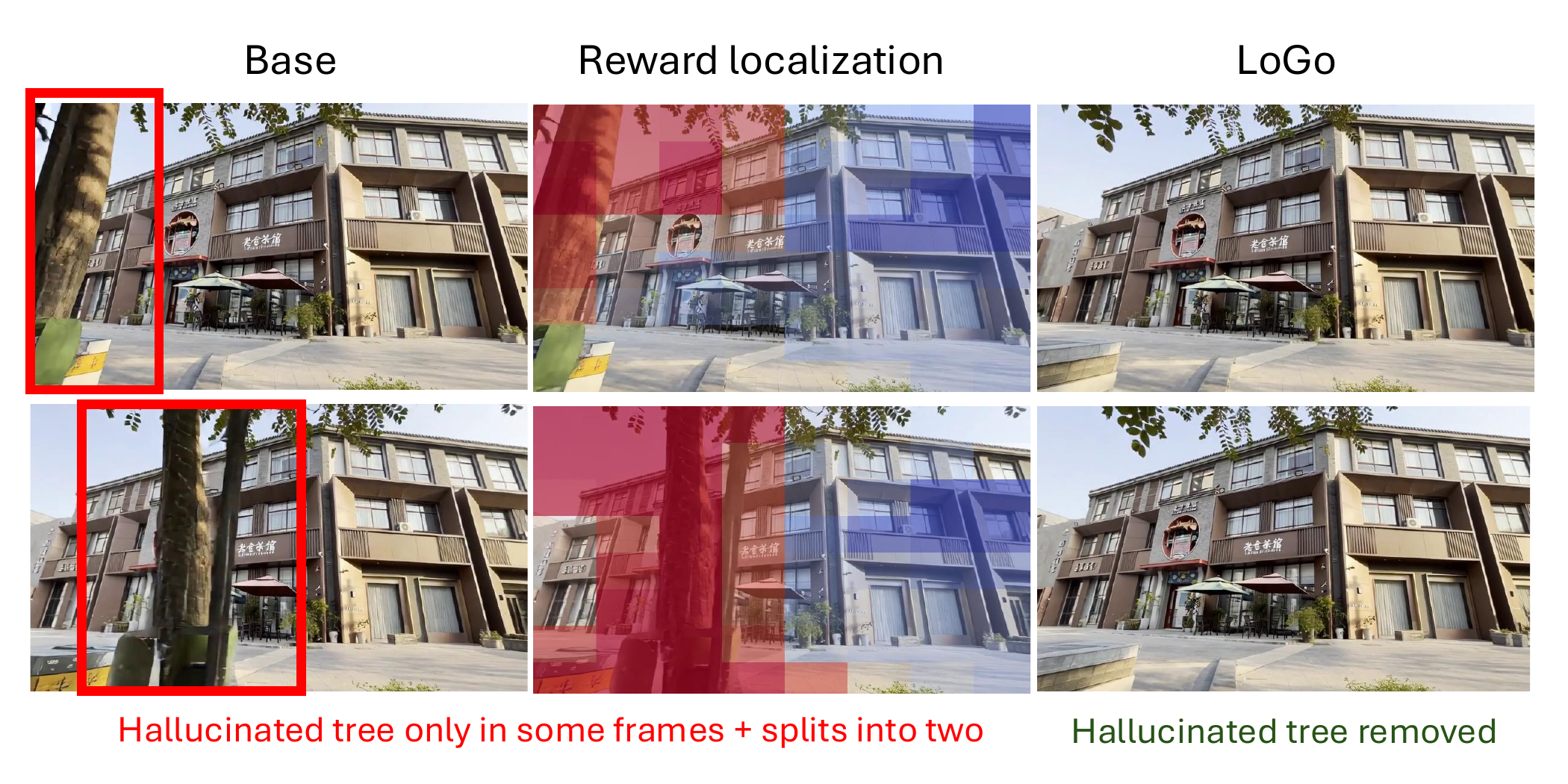}
\caption{Visualization of \method's local reward. The red-blue overlay is the normalized score per voxel projected back to image patches - blue means good and red means bad. In this example, a tree that did not exist in earlier is hallucinated in the middle of the video and splits into two. This is caught by \method's local reward component, which concentrates on the tree region, and subsequently fixed by post-training with \method.}
\label{fig:reward_visualize}
\end{figure}

We visualize voxel-localized credit assignment by back-projecting the score to image patches. The two keyframes show a tree non-existent in the start of the video hallucinated in the middle of the frame and splits into two parts. This inconsistency is caught by the local reward of \method -- the red (penalty) area concentrates around the tree. Training with \method removes the hallucinated tree.

\subsubsection{Voxel granularity ablation.}
\label{sec:additional_voxel}
%% -------- FIGURE -------- %%
\begin{figure}
  \centering
  \includegraphics[width=0.6\textwidth]{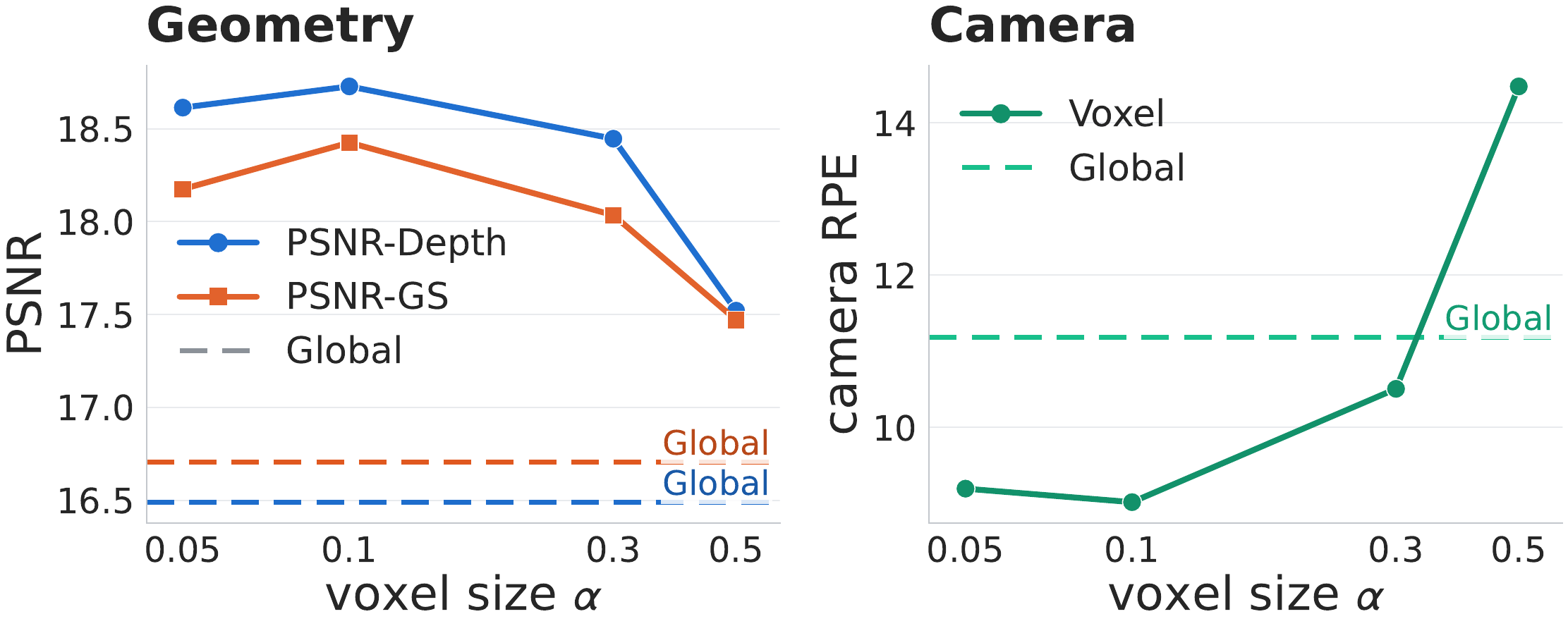}
\caption{Effect of voxel size on geometry metric and camera following metrics. Both improve with finer voxel size up till 0.1$\times$ P90 scene depth.}
\label{fig:voxel_size}
\end{figure}

%% ---------------- %%
We further ablate voxel granularity to understand what leads to the best credit assignment performance. Figure~\ref{fig:voxel_size} shows the effect of voxel granularity on geometry and camera control metrics. Since our scenes comprise of indoor and outdoor scenes of diverse scales, we define voxel size as a multiple of 90-percentile (P90) scene depth. Comparing voxel sizes of 0.05$\times$ to 0.7$\times$ P90 scene depth, we find that up to 0.1$\times$ P90 scene depth, finer granularity improves both 3D consistency geometry metrics (depth reprojection PSNR) and camera control. Finer-grained credit assignment intuitively helps improve the 3D objective better. Regarding camera, we hypothesize that penalizing a large voxel can incentivize the model to just turn the camera less to ``see" less of the penalized voxel, whereas fine-grained penalty does not provide this incentive: turning camera less could avoid the penalized small voxel, but also potentially other good voxels. Voxels that are too fine-grained no longer brings much improvement, since most inconsistencies can already be captured by the 0.1 granularity.

\end{document}